\documentclass[11pt]{article}

\usepackage[preprint]{acl}

\usepackage{times}
\usepackage{latexsym}
\usepackage[T1]{fontenc}
\usepackage[utf8]{inputenc}
\usepackage{inconsolata}
\usepackage{microtype}
\usepackage{graphicx}

\definecolor{darkblue}{rgb}{0, 0, 0.5}
\hypersetup{colorlinks=true, citecolor=darkblue, linkcolor=darkblue, urlcolor=darkblue}

\usepackage{hyperref}
\usepackage{booktabs}

\usepackage[colorinlistoftodos]{todonotes}

\usepackage{multirow}

\usepackage{pgf}
\usepackage{import}
\usepackage{mathptmx}

\usepackage{glossaries-extra}

\setabbreviationstyle[acronym]{long-short}
\glsdisablehyper

\newacronym[
  longplural={large language models},
  shortplural={LLMs}
]{llm}{LLM}{large language model}

\newacronym{tcp}{TCP}{Temporal Constraint-based Planning}
\newacronym{tot}{ToT}{Test of Time}
\newacronym{grpo}{GRPO}{Group Relative Policy Optimization}
\newacronym{auc}{AUC}{area under the curve}
\newacronym{delean}{DeLeAn}{Demand-Level-Annotation}
\newacronym{pca}{PCA}{principal component analysis}

\newacronym{qll}{QLl}{Logical Reasoning}
\newacronym{qlq}{QLq}{Quantitative Reasoning}
\newacronym{cee}{CEe}{Verbal Expression}
\newacronym{cec}{CEc}{Verbal Comprehension}
\newacronym{cl}{CL}{Conceptualisation, Learning and Abstraction}
\newacronym{knn}{KNn}{Knowledge of Natural Sciences}
\newacronym{kns}{KNs}{Knowledge of Social Sciences}
\newacronym{knc}{KNc}{Customary Everyday Knowledge}
\newacronym{sns}{SNs}{Spatio-physical Reasoning}
\newacronym{at}{AT}{Atypicality}
\newacronym{knf}{KNf}{Knowledge of Formal Sciences}

\title{Keeping Time vs. Understanding Time: Neither Alone Explains Temporal Reasoning in LLMs}

\author{
  \textbf{Auss Abbood\textsuperscript{1}},
  \textbf{Fernando Martínez-Plumed\textsuperscript{2}},
  \textbf{José Hernández-Orallo\textsuperscript{2,3}},
  \textbf{Nigel Collier\textsuperscript{1}} \\
  {\normalfont\textsuperscript{1}Language Technology Lab, University of Cambridge, UK} \\
  {\normalfont\textsuperscript{2}Valencian Research Institute for Artificial Intelligence (VRAIN)} \\
  {\normalfont Universitat Politècnica de València, Spain} \\
  {\normalfont\textsuperscript{3}Leverhulme Centre for the Future of Intelligence, University of Cambridge, UK} \\
  {\normalfont\small\texttt{\{aa2613,nhc30,jh2135\}@cam.ac.uk}, \texttt{fmartinez@dsic.upv.es}, \texttt{jorallo@upv.es}}
}

\hypersetup{
  pdftitle={Keeping Time vs. Understanding Time: Neither Alone Explains Temporal Reasoning in LLMs},
  pdfauthor={Auss Abbood, Fernando Martínez-Plumed, José Hernández-Orallo, Nigel Collier}
}

\begin{document}

\maketitle

\begin{abstract}
    Temporal reasoning is widely assumed to decompose into two modes: \emph{temporal arithmetic} (date and time calculations), and \emph{temporal semantics}  (understanding temporal and causal relations). This dichotomy shapes how benchmarks are designed, how models are trained, and how progress is measured, yet its empirical validity has never been directly tested. We provide two lines of evidence that challenge it.
    First, we test whether tasks labelled as arithmetic or semantic by the benchmarks support transferable gains within their purported mode. Finetuning on these subsets yields predominantly dataset-specific gains ($+6.8$ to $+23.1$ percentage points), while transfer to a different dataset in the same mode ranges from $-3.2$ to $+1.1$ percentage points. Across the two benchmark families used in the transfer analysis, this pattern provides little evidence of coherent, transferable temporal abilities.
    Second, using cognitive-demand-based model evaluation, we examine which capabilities actually drive performance on these benchmarks. We find that quantitative and logical reasoning, the assumed proxies for temporal arithmetic and temporal semantics, are not the primary drivers; knowledge is.
    Together, these findings show that current temporal reasoning benchmarks do not cleanly realise the arithmetic-semantics distinction and instead largely reward models for \emph{remembering} how time typically works.
\end{abstract}

\section{Introduction}

Which cognitive abilities are required to reason about time? Prior work on \glspl{llm} commonly decomposes temporal reasoning into two modes: \emph{temporal arithmetic}, involving calculations over times and dates, and \emph{temporal semantics}, involving the interpretation of before–after relations and temporal causality~\citep{wangTRAMBenchmarkingTemporal2024, chuTimeBenchComprehensiveEvaluation2024, fatemiTestTimeBenchmark2025, islakogluChronoSenseExploringTemporal2025}. Both require an understanding of time, whereas temporal arithmetic additionally requires mathematical capabilities. This distinction has become a common organising principle for benchmark design, model training, and performance reporting~\citep{wangTRAMBenchmarkingTemporal2024, chuTimeBenchComprehensiveEvaluation2024, suTimoBetterTemporal2024, fatemiTestTimeBenchmark2025}. However, it remains unclear whether it reflects meaningful differences in the cognitive demands tested by current benchmarks.

This question matters because benchmarks may not measure what they claim to measure due to poor construct validity~\citep{beanMeasuringWhatMatters2025}. As benchmarks become more complex, the true bottleneck may diverge from the target ability named in the task description~\citep{rameshkumarReasoningModelsReason2025,varelaRethinkingIllusionThinking2026}. 
For example, success on a temporal arithmetic question may depend more on numerical calculation than on temporal understanding, whereas success on a temporal semantics question may depend more on language comprehension or memorised world knowledge. The arithmetic/semantics distinction may therefore classify tasks by their surface form and not by the abilities that actually drive performance.

We test this assumption in two ways. First, we ask whether the arithmetic/semantics distinction predicts transfer across benchmarks. We train \glspl{llm} on temporal reasoning datasets and evaluate whether performance improvements generalise to unseen datasets, both within and across the two modes. If each mode reflects a stable capability with shared reasoning patterns, transfer should be strong between datasets in the same mode~\citep{shaoDeepSeekMathPushingLimits2024, huBreakingBarriersReinforcement2026, ionovCallRewardRepeat2026}. Instead, we find that improvements remain highly localised to the training data and barely generalise to unseen data in the same purported mode.

Second, we test whether the arithmetic/semantics split reflects differences in cognitive demands. Using the \gls{delean} rubrics~\citep{zhouGeneralScalesUnlock2026}, we score each benchmark item on 18 general cognitive dimensions. We compare arithmetic and semantic datasets across all \gls{delean} dimensions, but find no clear separation between the two modes. We then focus on the \gls{qlq} and \gls{qll} cognitive demands, which are most naturally aligned with temporal arithmetic~\citep{suTimoBetterTemporal2024, fatemiTestTimeBenchmark2025} and temporal semantics~\citep{xiongLargeLanguageModels2024, tangLTLBenchBenchmarksEvaluating2025}, respectively. We train models on data with relatively high demands for each of the two dimensions and test whether this produces larger gains in the corresponding mode. It does not. Instead, we find that tasks depend more strongly on the cognitive demands for knowledge.

Taken together, these results suggest that temporal reasoning benchmarks are better characterised as knowledge-heavy tasks than as tests of distinct arithmetic or semantic capabilities, with performance depending more on learned regularities and background knowledge than on explicit reasoning over temporal quantities or relations.

\section{Background}

\textbf{Temporal reasoning} benchmarks commonly distinguish tasks requiring semantic temporal understanding from those that also require mathematical operations. For example, \citet{suTimoBetterTemporal2024} note that 19 of the 38 tasks in TRAM~\citep{wangTRAMBenchmarkingTemporal2024} involve mathematical reasoning. TimeBench~\citep{chuTimeBenchComprehensiveEvaluation2024}, which covers 10 temporal reasoning benchmarks, distinguishes symbolic tasks such as date arithmetic from those requiring only semantic temporal understanding, while \gls{tot}~\citep{fatemiTestTimeBenchmark2025} explicitly has an Arithmetic and a Semantic subset. Across these efforts, the arithmetic/semantic split is treated as a meaningful axis of variation, yet none tests whether it reflects separable underlying abilities.

However, growing evidence questions the \textbf{construct validity} of LLM benchmarks---the extent to which they measure the abilities they claim to measure. Reviewing 445 benchmarks from major NLP and ML venues, \citet{beanMeasuringWhatMatters2025} find that many studies do not establish a clear link between the ability named by a benchmark and the abilities required to perform well on it. \citet{alaaPositionMedicalLarge2025} identify a similar gap in medicine, where high scores on exam-style benchmarks do not reliably predict performance on real clinical data. More broadly, \citet{rameshkumarReasoningModelsReason2025} and \citet{varelaRethinkingIllusionThinking2026} show that the main performance bottleneck can diverge from the capability named in the task description as tasks become more complex. These findings suggest that benchmark categories require empirical validation. We apply this principle to temporal reasoning by testing whether the arithmetic/semantic labels of existing benchmarks predict transfer behaviour and cognitive abilities.

If temporal arithmetic and temporal semantics reflect genuinely distinct and coherent abilities, then improvements in one should \textbf{transfer}, at minimum, to other tasks within the same mode. This logic follows a standard rationale in the reasoning literature: training in a narrow 
domain should generalise to tasks that draw on the same underlying capability. \citet{shaoDeepSeekMathPushingLimits2024} and \citet{guoDeepSeekR1IncentivizesReasoning2025} show that reinforcement learning on mathematical benchmarks reliably transfers to other mathematical tasks and even to non-mathematical reasoning benchmarks, confirming that fine-tuning \emph{can} yield cross-benchmark transfer when the underlying capability is shared. Whether the same holds for temporal reasoning 
remains untested. \citet{xiongLargeLanguageModels2024} show that fine-tuning on temporal data can improve temporal reasoning, but do not test cross-mode transfer, and recent work suggests that transfer is not guaranteed across all domains~\citep{huBreakingBarriersReinforcement2026, huanDoesMathReasoning2026}. We use fine-tuning 
transfer as a direct probe of the arithmetic/semantic distinction: if gains transfer within mode, the distinction has empirical grounding; if they do not, it lacks support as a description of separable abilities.

\section{Experimental setup}

\subsection{Datasets}

We use nine temporal reasoning datasets from eight benchmarks, counting ToT Arithmetic and Semantic separately (Table~\ref{tab:datasets}). For the demand-profile analyses and item-level experiments, TCP is further divided into Long and Short subsets, yielding ten subsets. These datasets span a broad range of temporal phenomena, task formats, and dataset construction methods. This variation allows us to test whether the arithmetic/semantic distinction generalises beyond a particular benchmark or task format. We report training, validation, and test splits for each dataset partition. The size of each split is reported in the Appendix (Tables~\ref{tab:base-dataset-cardinality} and~\ref{tab:delean-data-cardinality}).

\begin{table*}
    \centering
    \small   
    \begin{tabular}{lll}
        \toprule
        \textbf{Dataset}                          & \textbf{Format} & \textbf{Temporal phenomena}                          \\
        \midrule
        TCP  \citep{dingTCPBenchmarkTemporal2025}                & Free-form       & Constraint-based planning                            \\
        ToT Arithmetic  \citep{fatemiTestTimeBenchmark2025}  & Free-form       & Addition, duration, scheduling, time zones           \\
        ToT Semantic  \citep{fatemiTestTimeBenchmark2025}    & Free-form       & Event relationships, before/after, timelines         \\
        TRAM  \citep{wangTRAMBenchmarkingTemporal2024}              & Multiple-choice & Order, arithmetic, frequency, duration, NLI          \\
        TempReason  \citep{tanBenchmarkingImprovingTemporal2023}      & Free-form       & Temporal relation extraction                         \\
        TimeQA  \citep{chenDatasetAnsweringTimeSensitive2021}         & Free-form       & Time-sensitive question answering                    \\
        MenatQA  \citep{weiMenatQANewDataset2023}         & Free-form       & Counterfactual, shuffled context, varying time scope \\
        Date Arithmetic  \citep{tanBenchmarkingImprovingTemporal2023} & Free-form       & Date calculation                                     \\
        TimeDial  \citep{qinTIMEDIALTemporalCommonsense2021}       & Multiple-choice & Temporal common sense in dialogue                    \\
        \bottomrule
    \end{tabular}
    \caption{Overview of the nine temporal reasoning datasets from eight benchmarks. ToT Arithmetic and Semantic are counted separately.}\label{tab:datasets}
\end{table*}

\subsubsection{Cognitive capability labelling}\label{sec:delean}

We annotate each benchmark item \(X_j\) using the \gls{delean} rubric of \citet{zhouGeneralScalesUnlock2026}. GPT-4o assigns an annotation vector 
\(\mathbf{x}^{(j)}=(\vec{x}_1^{(j)},\ldots,x_{18}^{(j)})\in{\{0,\ldots,5^+}\}^{18}\), where \(x_i^{(j)}\) is the item's demand score on cognitive dimension \(i\), with higher values indicating greater demand. The dimensions include knowledge, metacognition, attention, comprehension, \textbf{\glsxtrfull{qlq}}, and \textbf{\glsxtrfull{qll}} (full definitions in Appendix~\ref{app:delean}).

The \gls{delean} rubrics build on established cognitive taxonomies and demonstrate good agreement among human raters and between the human consensus ratings and GPT-4o annotations. Five of the nine datasets in our study (Tab.~\ref{tab:datasets}) were included in the original evaluation of the framework: Date Arithmetic, MenatQA, TempReason, TimeDial, and TimeQA. Nevertheless, we confirm label reliability by relabelling data with Gemini 3.1 Flash Lite\footnote{On the \href{https://artificialanalysis.ai/models/comparisons/gemini-3-1-flash-lite-preview-vs-gpt-4o}{Artificial Analysis Intelligence Index} (v4.3.2), Gemini scores 16 versus GPT-4o's estimated 8 (November 2024).} and comparing its labels with GPT-4o’s, using within-group interrater reliability~\citep{jamesEstimatingWithingroupInterrater1984} as in the original study, but also explore ordinal Krippendorff's alpha~\citep{hayesAnsweringCallStandard2007}. More details are in the Appendix~\ref{app:label reliability}.

For the purposes of this study, \gls{qlq} and \gls{qll} are the most relevant dimensions, as they map most directly onto the conventional arithmetic/semantic distinction. \Gls{qlq} captures the numerical demands associated with temporal arithmetic~\citep{suTimoBetterTemporal2024, fatemiTestTimeBenchmark2025}, while \Gls{qll} captures the relational and causal reasoning demands associated with temporal semantics~\citep{wangTRAMBenchmarkingTemporal2024, xiongLargeLanguageModels2024}. By comparing demand levels on these two dimensions at the item level, we can test whether the arithmetic/semantic distinction corresponds to a genuine difference in cognitive requirements, independently of the labels assigned by benchmark designers.

\subsubsection{Operationalising the arithmetic/semantic distinction}\label{sec:ari-sem-distinction}

We use fine-tuning transfer (see~\ref{sec:ft}) between temporal arithmetic and semantics as a probe of their separability. If both modes reflect distinct abilities, then training on one should improve performance on tasks from the same mode, and not on tasks from the other. We use two complementary operationalisations to partition the data into these modes.

The first is a \emph{dataset-level} operationalisation based on existing mode labels. We restrict this controlled transfer analysis to \gls{tot} and TRAM, as these are the only benchmarks that provide explicit or reconstructable arithmetic/semantic partitions, minimising confounds from differences in dataset construction, such as annotation assumptions, or task format. \gls{tot} provides an explicit split: its Arithmetic subset covers time/date arithmetic, duration, scheduling, and time zone conversion tasks, while its Semantic subset targets event--event/time, before/after, timeline, and first/last relations. TRAM does not provide such a binary partition. Instead, we follow \citeposs{suTimoBetterTemporal2024} classification of TRAM subtasks into \emph{math-time tasks} (equivalent to arithmetic mode) and \emph{pure-time tasks} (equivalent to semantic mode) (see Appendix~\ref{app:tram} for details).

The second is an \emph{item-level} operationalisation based on \gls{delean} annotations (Section~\ref{sec:delean}). We partition items by their \gls{qlq} and \gls{qll} demands to construct training sets intended to probe both temporal reasoning modes. Specifically, we filter all datasets to items where demand for \gls{qlq} exceeds that for \gls{qll}, and vice versa, yielding six training conditions. These include two zero-demand controls (Tab.~\ref{tab:cognitive-training-conditions}). It is worth noting that \gls{qlq} and \gls{qll} are rarely the most demanding dimensions in our data. Out of 8,530 training items, \gls{qlq} is the dominant dimension in only 15 items (1,288 including ties), and \gls{qll} in only 1 item (771 including ties). We will discuss the limited role these dimensions seem to play in current benchmarks later.

\begin{table}
    \centering
    \small
    \begin{tabular}{
        l
        r@{\hspace{0.35em}}
        c@{\hspace{0.35em}}
        l
        r
    }
        \toprule
        \textbf{Condition} & \multicolumn{3}{c}{\textbf{Item filter}} & \textbf{$n$} \\
        \midrule
        Control      & $\mathrm{QLl}$ & $=$ & $0$              & 938 \\
                  & $\mathrm{QLq}$ & $=$ & $0$              & 2,425 \\
        \addlinespace
        QLq-dominant & $\mathrm{QLq}$ & $=$ & $\mathrm{QLl}+1$ & 2,479 \\
                     & $\mathrm{QLq}$ & $=$ & $\mathrm{QLl}+2$ & 247 \\
        \addlinespace
        QLl-dominant & $\mathrm{QLl}$ & $=$ & $\mathrm{QLq}+1$ & 1,826 \\
                     & $\mathrm{QLl}$ & $=$ & $\mathrm{QLq}+2$ & 54 \\
        \bottomrule
    \end{tabular}
    \caption{Item-level operationalisation based on relative
    \gls{qlq} and \gls{qll} demand.}
    \label{tab:cognitive-training-conditions}
\end{table}

\subsubsection{Non-temporal QL-heavy training}
\label{sec:ql-heavy-training}

QL demand is low in the temporal data
(\(\operatorname{mean}(\max(\mathrm{QLq},\mathrm{QLl}))=1.57\);
Fig.~\ref{fig:benchmark_heatmap}). To test whether a stronger QL signal
improves temporal arithmetic and semantics, we select non-temporal examples from
ChemLLMBench~\citep{guoWhatCanLarge2023}, MMLU-Pro~\citep{wangMMLUproMoreRobust2024},
AGIEval~\citep{zhongAGIEvalHumancentricBenchmark2024},
TruthQuest~\citep{mondorfLiarLiarLogical2024},
OmniMath~\citep{gaoOmniMATHUniversalOlympiad2025}, and
LiveBench~\citep{whiteLiveBenchChallengingContaminationlimited2025}.
Filtering for \(\mathrm{QL}=\max(\mathrm{QLq},\mathrm{QLl})\ge3\) and
\(\mathrm{QL}>\max_{d\in D_{\setminus\mathrm{QL}}}d\) yields 428 items
with a mean QL demand of 4.13.

\subsection{Evaluation}
\label{sec:evaluation}

We use accuracy for evaluation and additionally normalise temporal expressions before scoring, following \citet{abboodTimeRevisitExact2025}. The expected answer type in the dataset guides the normalisation of the model’s response. E.g., ``11 Aug. 2023'' and ``11/08/2023'' are then treated as equivalent. This prevents surface-level differences in temporal formatting from being counted as reasoning failures (details in Appendix~\ref{app:metrics}).

Aggregate accuracy, however, does not isolate performance on any single cognitive demand dimension, since each item carries a multidimensional \gls{delean} label. To disentangle the contribution of individual dimensions, we  construct \emph{subject characteristic curves} using dominant slices. For a dominant dimension \(i\)  and level \(l\), we retain only items satisfying \(x_{i}^{(j)} = l\) and \(x_{i^{'}}^{(j)} \leq l\) for all \(i^{'} \neq i\), i.e., items for which dimension \(i\) is the dominant demand. Accuracy is averaged within each level and a two-parameter logistic curve is fitted to the resulting five points, using the weighting scheme of \citet{zhouGeneralScalesUnlock2026}. We report the \gls{auc} of the fitted curve as the model's ability on that dimension.

Four of the 18 dimensions are excluded from this analysis due to insufficient coverage in our data. \citeauthor{zhouGeneralScalesUnlock2026} suggests excluding dimensions that have a range narrower than 1--3 or have fewer than 30 items. \gls{sns} and \gls{cee} span only demand levels 1 and 2, too narrow a range for meaningful curve fitting. \gls{cl} and \gls{knn} yield only 9 and 25 items, respectively, across all dominant slices, making estimates unreliable.

\subsection{Fine-tuning} \label{sec:ft}

Our main method for investigating the separability between temporal arithmetic and temporal semantics is to fine-tune models on data for each mode and test the transfer of performance to the corresponding mode.

\paragraph{GRPO.} Temporal reasoning tasks can require multi-step reasoning. We initially attempted supervised fine-tuning on reasoning traces from a larger teacher model~\citep{hsiehDistillingStepbyStepOutperforming2023}, without success. We attribute this to a mismatch between the teacher's and student's response style~\citep{liSmallModelsStruggle2025}. Instead, we use \gls{grpo}~\citep{guoDeepSeekR1IncentivizesReasoning2025}, which sidesteps this issue by optimising directly on self-generated reasoning traces. To ensure comparability across experiments despite substantial variation in dataset sizes, we fix the compute budget to 400 optimisation steps with a batch size of 16. We use 8 generations per rollout for dataset-level operationalisation and 16 for the item-level operationalisation, due to training instabilities with 8 rollouts (details in Appendix~\ref{app:training-settings}).

\paragraph{Reward.} We define two reward settings to probe the effect of training from complementary directions. In the \emph{accuracy} setting, the reward is 1 if the model's response is correct and 0 otherwise, determined by the semantic equivalence metric described in Section~\ref{sec:evaluation}. In the \emph{one-minus-accuracy} setting, we invert this signal to explore a form of targeted unlearning similar to \citet{zaradoukasReinforcementUnlearningGroup2026}. The model receives a reward of 1 only if its answer is incorrect but correctly formatted, 0.5 if the answer is both correct and correctly formatted (for richer feedback), and 0 otherwise. The format requirement mitigates reward hacking~\citep{skalseDefiningCharacterizingReward2022} by preventing models from maximising reward through degenerate outputs. If temporal arithmetic and semantics are separable, unlearning one mode should degrade performance primarily on the same mode, leaving the other largely intact. For experiments using the item-level operationalisation, we additionally scale the reward by \( 1 + \text{QLq} + \text{QLl} \), so that items with higher quantitative or logical demand contribute a stronger optimisation signal. The goal is to offset the low-resolution binary reward signal, which initially provided limited learning signal across many samples.

\paragraph{Models.} We run our experiments with six models from three families and two size ranges, ensuring that our findings are not an artefact of a single architecture or scale. Specifically, we use Qwen3-4B, Qwen3-8B~\citep{yangQwen3TechnicalReport2025}, Llama-3.2-3B-Instruct, Llama-3.1-8B-Instruct~\citep{grattafioriLlama3Herd2024}, gemma-3-4B-it, and gemma-3-12B-it~\citep{gemmateamGemma3Technical2025}. Qwen models are run in non-thinking mode for comparability with the other models. During evaluation, we allow up to 4,000 output tokens and set the temperature to 0 for deterministic decoding.

\section{Results}

\subsection{Cognitive demand profiles do not reflect the temporal reasoning dichotomy}\label{sec:cognitive-demand}

If the arithmetic/semantic distinction reflects a genuine structural property of temporal reasoning, it should be recoverable through the data's cognitive demands. Specifically, items labelled as temporal arithmetic should exhibit elevated \gls{qlq} demands relative to \gls{qll}, while items labelled as temporal semantics should show the reverse. We test this prediction in two ways: first by examining the joint distribution of \gls{qlq} and \gls{qll} across all benchmarks, and then by examining whether the full 18-dimensional demand profiles cluster by mode.

\begin{figure}
    \centering
    \input{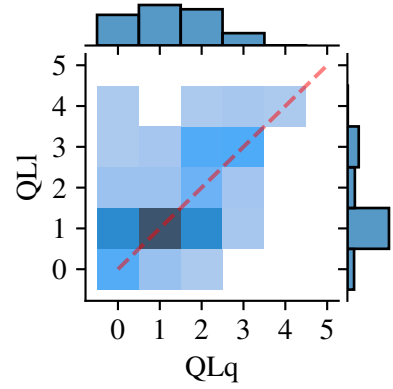}
    \caption{Joint distribution of \gls{qlq} and \gls{qll} demand levels across all datasets.} \label{fig:qlq-qll-dist}
 \end{figure}

\paragraph{QLq and QLl do not diverge.} Fig.~\ref{fig:qlq-qll-dist} shows that the joint distribution of \gls{qlq} and \gls{qll} demand levels across all benchmark items is mostly similar. Items primarily concentrate near the diagonal, with the modal demand level at 1 for both dimensions. \Gls{qll} tends to exceed \gls{qlq} slightly more often and reaches higher levels, but the two dimensions are broadly correlated rather than orthogonal.

This pattern persists even in \gls{tot}, which is explicitly designed to separate arithmetic from semantic tasks. \Gls{tot} Semantic has 679 items where \(\text{QLq} \ge 3\), while the Arithmetic subset has only 41 items, making the Semantic subset quantitatively \emph{more} \gls{qlq} demanding. This is partly explained by two of the eight Semantic question types (\emph{RelationDuration} and \emph{NumEventsInTimeInterval}) which require arithmetic operations (see Appendix~\ref{app:tot-semantic} for a breakdown of \gls{qlq} and \gls{qll} scores per question type). The full distribution of demand levels for both \gls{tot} subsets is depicted in the Appendix (Fig.~\ref{fig:benchmark_pies}).

\begin{figure}
    \centering 
    \import{figures/}{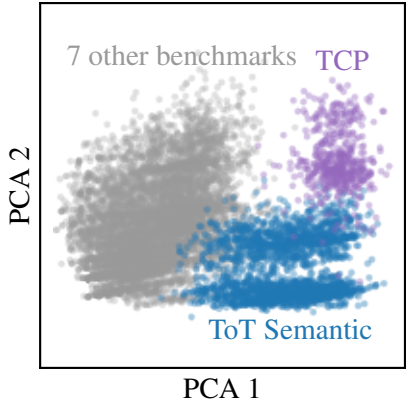}
    \caption{PCA of the full 18-dimensional \gls{delean} profiles (PC1: 40.5\%, PC2: 15.5\%; 56.0\% variance explained). TCP and ToT Semantic are annotated; the remaining seven datasets are shown in grey. TCP Long and Short share the TCP label.} \label{fig:pca_2d}
\end{figure}

\paragraph{\Gls{delean} separates benchmarks by overall demand, not by mode.} Because \gls{qlq} and \gls{qll} alone do not distinguish the two purported modes, we ask whether the distinction emerges when all 18 cognitive dimensions are considered. We therefore apply \gls{pca} to the DeLeAn-labelled items (Fig.~\ref{fig:pca_2d}). The first two principal components' explained variance is PC1: 40.5\% and PC2: 15.5\%. If the arithmetic/semantics distinction were a major organising principle, we would expect items from both modes to occupy different regions of this projection. Instead, the dominant structure appears to reflect overall cognitive demand\footnote{Overall demand is calculated as the average demand level across all 18 dimensions; see Fig.~\ref{fig:benchmark_heatmap} in Appendix~\ref{app:benchmark-profiles}}: \gls{tcp} and \gls{tot} Semantic, the two most demanding datasets in this study, each form a distinct cluster. The remaining benchmarks show no mode-consistent grouping. In short, the leading principal components capture differences in overall benchmark demand but reveal no clustering that supports the arithmetic/semantics distinction.

\subsection{Performance gains are dataset-specific and do not transfer across modes}\label{sec:transfer}

The cognitive demand analysis above examines what benchmarks \emph{require}; we now test what models actually \emph{learn}. If temporal arithmetic and temporal semantics are separable abilities,  then fine-tuning on one mode should improve performance on other datasets from the same mode, and not on datasets from another mode. 

\begin{figure*}[t]
\centering
\import{figures/}{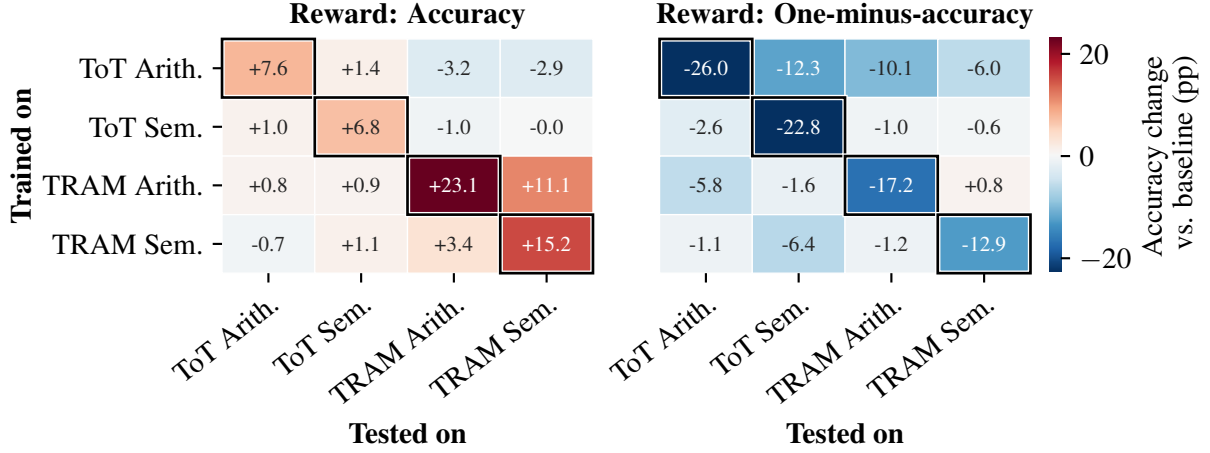}
\caption{Average relative change in accuracy percentage points between the untrained model (baseline) and the trained model. Results averaged over six models from three model families and two sizes. Results shown for the accuracy setting (reward given for correct answers) and one-minus-accuracy setting (reward given for an incorrect but correctly formatted answer).}\label{fig:accuracy_delta_matrix}
\end{figure*}

To test transfer, we treat the arithmetic and semantic portions of each benchmark as separate subsets, yielding four dataset-mode combinations: ToT-Arithmetic, ToT-Semantic, TRAM-Arithmetic, and TRAM-Semantic. We fine-tune six models separately on each combination and evaluate resulting variants on all four test sets. This design allows us to test transfers across datasets within and across temporal reasoning modes. For each training–test combination, Fig.~\ref{fig:accuracy_delta_matrix} reports the change in accuracy relative to the corresponding untrained model, averaged across the six models.

\paragraph{Within- and cross-mode transfer fails.}
In the accuracy reward setting, in-domain gains are substantial and consistent (diagonal in Fig.~\ref{fig:accuracy_delta_matrix}, left). These gains, however, do not transfer across benchmarks, even when the source and target task share a mode. If the arithmetic/semantic distinction captured coherent, transferable abilities, within-mode transfer should be positive and reliable\footnote{This is not inherent to the training method: \gls{grpo} yields reliable cross-benchmark
transfer in mathematical reasoning~\citep{shaoDeepSeekMathPushingLimits2024, guoDeepSeekR1IncentivizesReasoning2025, huBreakingBarriersReinforcement2026}.}. Instead, off-diagonal transfer is essentially absent, with one partial exception: training on TRAM arithmetic improves TRAM semantic by 11.1~pp (though the within-dataset gain is far larger, 23.1~pp), and training on TRAM semantic improves TRAM arithmetic by 3.4~pp (with a 15.2~pp gain on TRAM semantic itself). Training on \gls{tot} arithmetic, by contrast, degrades TRAM performance ($-3.2$ and $-2.9$~pp). Inspecting model responses shows within-TRAM transfer is largely a formatting effect. The two TRAM modes share the prompt template and a single multiple-choice-letter answer space. Thus, improving answer formatting in one mode recovers otherwise-unparseable responses in the other. The effect does not replicate for \gls{tot}, whose two modes differ in both content and answer format (structured JSON vs.\ event-identifier sets).

\paragraph{Unlearning confirms the pattern.} The one-minus-accuracy experiments (Fig.~\ref{fig:accuracy_delta_matrix}, right) provide a complementary perspective. Performance degradation mirrors the localised, dataset-specific nature of accuracy gains. This effect is most pronounced for \gls{tot} arithmetic ($-26$ pp) and ToT Semantic ($-22.8$ pp). However, degradation spreads more broadly off-diagonal than improvement does. 

Notably, gains and losses are not symmetric across reward settings. For TRAM, accuracy training transfers arithmetic improvements to semantics more strongly than the reverse, whereas unlearning reverses this asymmetry. This suggests that the two TRAM subsets share surface-level regularities that are learned and unlearned asymmetrically, rather than drawing on a common underlying ability.

Overall, these results suggest that abilities learned during fine-tuning are highly localised to individual datasets rather than transferable improvements in temporal arithmetic or semantic reasoning. Per-model results (Appendix~\ref{app:per-model}, 
Tab.~\ref{tab:model-performance-deltas}) confirm that this pattern is consistent across all six models and not driven by individual outliers.

An exploratory experiment with Qwen3-32B produces smaller and mixed transfer effects, including a 3.7-point gain from TRAM Arithmetic to \gls{tot} Arithmetic, but does not establish a consistent mode-specific pattern (Appendix~\ref{app:larger_models}).

\subsection{Quantitative and logical reasoning are not the main drivers for performance}\label{sec:drivers}

The preceding transfer experiments use benchmark-defined labels to distinguish the two temporal-reasoning modes. We complement them with an item-level operationalisation by training on data filtered by relative \gls{qlq} and \gls{qll} demands (Sec.~\ref{sec:ari-sem-distinction}) and measuring changes in \gls{delean}-defined capabilities. This asks a finer-grained question: when trained on data with high quantitative or logical demands, do models improve most on those dimensions, as the arithmetic/semantic distinction predicts, or on other capabilities?

Accuracy for evaluation is insufficient here, because we are interested in changes across the full capability profile rather than end-task performance alone. We therefore compute DeLeAn-based subject characteristic curves as described in Section~\ref{sec:evaluation} and use the \gls{auc} across 14 retained dimensions as a summary measure of model ability on each dimension. Fig.~\ref{fig:heatmap_dw_oma_adele} reports the average change in \gls{auc} relative to baseline, averaged across all six models for both experimental settings. Per-model results are in the Appendix~\ref{app:per-model}.

\begin{figure*}[t]
    \centering
    (a) Reward: Accuracy\par
    \import{figures/}{heatmap_dw_adele.pgf}

    \vspace{0pt}
    \noindent\rule{0.9\linewidth}{0.1pt}
    \vspace{0pt}

    (b) Reward: One-minus-accuracy\par
    \import{figures/}{heatmap_oma_adele.pgf}

    \caption{Average absolute change in AUC between the untrained model (baseline) and the trained model. Results are averaged over six models from three model families and two sizes. Panel (a) shows the accuracy reward setting, and panel (b) shows the one-minus-accuracy reward setting.}
    \label{fig:heatmap_dw_oma_adele}
\end{figure*}

\paragraph{QLq and QLl's subordinate role in improving temporal reasoning.} The arithmetic and semantic modes contrast temporal reasoning that requires explicit calculation with temporal reasoning that does not. If true, we would expect \gls{qlq}-dominant training to primarily improve \gls{qlq} (and hence temporal arithmetic), and the equivalent for \gls{qll}-dominant training. This is not what we observe. Training on data where \(\text{QLl}=\text{QLq}+1\) yields the largest improvements for both \gls{qlq} and \gls{qll}, rather than selectively boosting \gls{qll}. The \(\text{QLl}=\text{QLq}+2\) condition produces weaker gains, possibly due to its small dataset size (54 items). \Gls{qlq}-dominant training, where \(\text{QLq}=\text{QLl}+\{1,2\}\), also does not preferentially improve \gls{qlq}. Because \gls{qlq} demands are generally lower than \gls{qll} demands across the benchmarks (Fig.~\ref{fig:benchmark_heatmap} in the Appendix), \gls{qlq}-dominant items may not provide a sufficiently challenging learning signal for that dimension.

\paragraph{A stronger QL signal does not improve temporal accuracy.} To test whether low QL demand in the temporal training data (mean demand of 1.57) explains the limited performance gains, we fine-tune models on QL-dominant, non-temporal data (mean demand of 4.13). Although \gls{qll} and \gls{qlq} both increase by 0.15 AUC---the largest capability gains among all dimensions---accuracy on the temporal mix data remains unchanged (mean $\Delta=-0.06$~pp; range $-1.23$ to $+0.55$; Appendix~\ref{app:ql-heavy}, Tables~\ref{tab:non-temporal-ql-fam-capability-auc},\ref{tab:non-temporal-ql-fam-temporal-mix-acc}). Thus, stronger QL capability is insufficient to improve temporal reasoning and QL is not the principal bottleneck in model performance.

\paragraph{Knowledge benefits the most.} The most revealing finding (Fig.~\ref{fig:heatmap_dw_oma_adele}, top) emerges when we examine which dimensions benefit most across all six training conditions. Gains in \gls{kns}, \gls{knc}, and \gls{cec} consistently exceed gains in \gls{qll} and \gls{qlq} (often by a wide margin). This pattern holds regardless of whether the training data was selected for its quantitative or logical demands. In other words, even when we deliberately select items that emphasise arithmetic or logical reasoning, models primarily improve on \gls{knc}, \gls{kns} and \gls{cec}.  The one-minus-accuracy results (Fig.~\ref{fig:heatmap_dw_oma_adele}, bottom) corroborate this from the reverse direction: the largest capability degradations occur for the \gls{kns} dimension, confirming that these are the abilities most exercised by current temporal benchmarks. The per-benchmark demand distributions (Appendix~\ref{app:benchmark-profiles}, Fig.~\ref{fig:benchmark_pies}) provide further support. Most benchmarks concentrate their \gls{qlq} and \gls{qll} demands at levels 0--2, with few items at the higher levels, which would be necessary to expect reasoning bottlenecks. 

Relabelling with Gemini largely reproduces the KN-over-QL pattern seen in Fig.~\ref{fig:heatmap_dw_oma_adele}, although \gls{kns} rather than \gls{knc} benefits most and \gls{cec} gains are smaller (details in Appendix~\ref{app:label reliability}).

These results suggest that logical and quantitative reasoning are not the primary drivers of performance on current temporal reasoning benchmarks. Training on examples selected for their logical or quantitative demands does affect those dimensions, but the dominant effect falls on knowledge dimensions. This finding reinforces the conclusion from Sections~\ref{sec:cognitive-demand} and~\ref{sec:transfer}: existing temporal reasoning benchmarks do not cleanly instantiate the arithmetic/semantics distinction. Instead, they largely measure models' capacity to retrieve and apply learned temporal regularities.

\section{Conclusions}

This study scrutinises whether temporal arithmetic and temporal semantics correspond to distinct reasoning abilities. Capability-aware analyses of the dataset and fine-tuning experiments show that they do not. Cognitive demand profiles fail to separate arithmetic from semantic benchmark items, despite supposed differences in required quantitative and logical reasoning. Furthermore, finetuning gains do not transfer between benchmarks of the same temporal reasoning mode. 

Capability changes after finetuning on temporal data show that \gls{delean} knowledge dimensions improve more than \gls{qlq} or \gls{qll}. Specialised QL-heavy training could raise QL capability, but still did not improve performance in either temporal reasoning modes. Our results suggest that current benchmarks test the retrieval and application of time-related regularities, not arithmetic or semantic abilities. The arithmetic/semantics taxonomy may describe task form, but it should not be treated as a salient decomposition of temporal reasoning. Future benchmarks should control for knowledge demands and validate proposed categories through capability-based analyses.

\section{Limitations}

Our experiments use models of up to 12 billion parameters. In one instance, we trained a larger model, which showed small transfer (improvement of $\approx 3$ percentage points) in the arithmetic mode. Future work could explore transfer dynamics in even larger models; however, the consistency of our findings across three families and two scales suggests that this pattern is robust. Furthermore, our transfer experiments are limited to \gls{tot} and TRAM, which are the only benchmarks with explicit or reconstructable arithmetic and semantic partitions. Extending this analysis to future benchmarks that provide more detailed task taxonomies would strengthen the generalisability of our findings.

\section*{Acknowledgments}

JHO's research is supported by OpenAI's grant to the ``AI Progress through the Lens of Predictable AI Ecosystems'' programme, which is based at the Leverhulme Centre for the Future of Intelligence at the University of Cambridge.
JHO's research has also been supported by Google's grant to the ``Evaluations and Benchmarks'' project, which is based at the Leverhulme Centre for the Future of Intelligence at the University of Cambridge.
AA is supported by the Economic and Social Research Council [ES/Y001788/1].

\bibliography{references}

\begin{thebibliography}{48}
\providecommand{\natexlab}[1]{#1}

\bibitem[{Abbood et~al.(2025)Abbood, Meng, and
  Collier}]{abboodTimeRevisitExact2025}
Auss Abbood, Zaiqiao Meng, and Nigel Collier. 2025.
\newblock \href {https://doi.org/10.18653/v1/2025.findings-emnlp.637} {Time to
  {Revisit} {Exact} {Match}}.
\newblock In \emph{Findings of the {Association} for {Computational}
  {Linguistics}: {EMNLP} 2025}, pages 11903--11926, Suzhou, China. Association
  for Computational Linguistics.

\bibitem[{Alaa et~al.(2025)Alaa, Hartvigsen, Golchini, Dutta, Dean, Raji, and
  Zack}]{alaaPositionMedicalLarge2025}
Ahmed Alaa, Thomas Hartvigsen, Niloufar Golchini, Shiladitya Dutta, Frances
  Dean, Inioluwa~Deborah Raji, and Travis Zack. 2025.
\newblock \href {https://proceedings.mlr.press/v267/alaa25a.html} {Position:
  {Medical} large language model benchmarks should prioritize construct
  validity}.
\newblock In \emph{Proceedings of the 42nd international conference on machine
  learning}, volume 267 of \emph{Proceedings of machine learning research},
  pages 80991--81004. PMLR.

\bibitem[{Ansel et~al.(2024)Ansel, Yang, He, and
  {others}}]{anselPyTorch2Faster2024}
Jason Ansel, Edward Yang, Horace He, and {others}. 2024.
\newblock \href {https://doi.org/10.1145/3620665.3640366} {{PyTorch} 2:
  {Faster} machine learning through dynamic python bytecode transformation and
  graph compilation}.
\newblock In \emph{Proceedings of the 29th {ACM} international conference on
  architectural support for programming languages and operating systems, volume
  2}, pages 929--947. Association for Computing Machinery.

\bibitem[{Bean et~al.(2025)Bean, Kearns, Romanou, Hafner, Mayne, Batzner,
  Foroutan~Eghlidi, Schmitz, Korgul, Batra, Deb, Beharry, Emde, Foster, Gausen,
  Grandury, Han, Hofmann, Ibrahim, Kim, Kirk, Lin, Liu, Luettgau, Magomere,
  Rystrøm, Sotnikova, Yang, Zhao, Bibi, Bosselut, Clark, Cohan, Foerster, Gal,
  Hale, Raji, Summerfield, Torr, Ududec, Rocher, and
  Mahdi}]{beanMeasuringWhatMatters2025}
Andrew~M. Bean, Ryan~Othniel Kearns, Angelika Romanou, Franziska~Sofia Hafner,
  Harry Mayne, Jan Batzner, Negar Foroutan~Eghlidi, Chris Schmitz, Karolina
  Korgul, Hunar Batra, Oishi Deb, Emma Beharry, Cornelius Emde, Thomas Foster,
  Anna Gausen, María Grandury, Sophia Han, Valentin Hofmann, Lujain Ibrahim,
  and 23 others. 2025.
\newblock \href
  {https://proceedings.neurips.cc/paper_files/paper/2025/file/1967e0fc3aa6cbbace562f5cb8e3954e-Paper-Datasets_and_Benchmarks_Track.pdf}
  {Measuring what matters: {Construct} validity in large language model
  benchmarks}.
\newblock In \emph{Advances in neural information processing systems},
  volume~38. Curran Associates, Inc.

\bibitem[{Chen et~al.(2021)Chen, Wang, and
  Wang}]{chenDatasetAnsweringTimeSensitive2021}
Wenhu Chen, Xinyi Wang, and William~Yang Wang. 2021.
\newblock \href
  {https://datasets-benchmarks-proceedings.neurips.cc/paper/2021/hash/1f0e3dad99908345f7439f8ffabdffc4-Abstract-round2.html}
  {A {Dataset} for {Answering} {Time}-{Sensitive} {Questions}}.
\newblock \emph{Proceedings of the Neural Information Processing Systems Track
  on Datasets and Benchmarks}, 1.

\bibitem[{Chu et~al.(2024)Chu, Chen, Chen, Yu, Wang, Liu, and
  Qin}]{chuTimeBenchComprehensiveEvaluation2024}
Zheng Chu, Jingchang Chen, Qianglong Chen, Weijiang Yu, Haotian Wang, Ming Liu,
  and Bing Qin. 2024.
\newblock \href {https://doi.org/10.18653/v1/2024.acl-long.66} {{TimeBench}:
  {A} {Comprehensive} {Evaluation} of {Temporal} {Reasoning} {Abilities} in
  {Large} {Language} {Models}}.
\newblock In \emph{Proceedings of the 62nd {Annual} {Meeting} of the
  {Association} for {Computational} {Linguistics} ({Volume} 1: {Long}
  {Papers})}, pages 1204--1228, Bangkok, Thailand. Association for
  Computational Linguistics.

\bibitem[{Ding et~al.(2025)Ding, Yan, Yuan, Hu, Lin, and
  Vlachos}]{dingTCPBenchmarkTemporal2025}
Zifeng Ding, Sikuan Yan, Moy Yuan, Xianglong Hu, Fangru Lin, and Andreas
  Vlachos. 2025.
\newblock \href {https://doi.org/10.18653/v1/2025.emnlp-main.1142} {{TCP}: a
  {Benchmark} for {Temporal} {Constraint}-{Based} {Planning}}.
\newblock In \emph{Proceedings of the 2025 {Conference} on {Empirical}
  {Methods} in {Natural} {Language} {Processing}}, pages 22452--22475, Suzhou,
  China. Association for Computational Linguistics.

\bibitem[{Fatemi et~al.(2025)Fatemi, Kazemi, Tsitsulin, Malkan, Yim, Palowitch,
  Seo, Halcrow, and Perozzi}]{fatemiTestTimeBenchmark2025}
Bahare Fatemi, Mehran Kazemi, Anton Tsitsulin, Karishma Malkan, Jinyeong Yim,
  John Palowitch, Sungyong Seo, Jonathan Halcrow, and Bryan Perozzi. 2025.
\newblock \href {https://openreview.net/forum?id=44CoQe6VCq} {Test of {Time}:
  {A} {Benchmark} for {Evaluating} {LLMs} on {Temporal} {Reasoning}}.
\newblock In \emph{The {Thirteenth} {International} {Conference} on {Learning}
  {Representations}}, Singapore.

\bibitem[{Gao et~al.(2025)Gao, Song, Yang, Cai, Miao, Dong, Li, Ma, Chen, Xu,
  Tang, Wang, Zan, Quan, Zhang, Sha, Zhang, Ren, Liu, and
  Chang}]{gaoOmniMATHUniversalOlympiad2025}
Bofei Gao, Feifan Song, Zhe Yang, Zefan Cai, Yibo Miao, Qingxiu Dong, Lei Li,
  Chenghao Ma, Liang Chen, Runxin Xu, Zhengyang Tang, Benyou Wang, Daoguang
  Zan, Shanghaoran Quan, Ge~Zhang, Lei Sha, Yichang Zhang, Xuancheng Ren,
  Tianyu Liu, and Baobao Chang. 2025.
\newblock \href {https://openreview.net/forum?id=yaqPf0KAlN} {Omni-{MATH}: a
  universal olympiad level mathematic benchmark for large language models}.
\newblock In \emph{The thirteenth international conference on learning
  representations}.

\bibitem[{{Gemma Team}(2025)}]{gemmateamGemma3Technical2025}
{Gemma Team}. 2025.
\newblock \href {https://arxiv.org/abs/2503.19786} {Gemma 3 technical report}.
\newblock ArXiv: 2503.19786 [cs.CL].

\bibitem[{Grattafiori et~al.(2024)Grattafiori, Dubey, Jauhri, Pandey, Kadian,
  Al-Dahle, Letman, Mathur, Schelten, Vaughan, Yang, Fan, Goyal, Hartshorn,
  Yang, Mitra, Sravankumar, Korenev, Hinsvark, Rao, Zhang, Rodriguez,
  Gregerson, Spataru, Roziere, Biron, Tang, Chern, Caucheteux, Nayak, Bi,
  Marra, McConnell, Keller, Touret, Wu, Wong, Ferrer, Nikolaidis, Allonsius,
  Song, Pintz, Livshits, Wyatt, Esiobu, Choudhary, Mahajan, Garcia-Olano,
  Perino, Hupkes, Lakomkin, AlBadawy, Lobanova, Dinan, Smith, Radenovic,
  Guzmán, Zhang, Synnaeve, Lee, Anderson, Thattai, Nail, Mialon, Pang,
  Cucurell, Nguyen, Korevaar, Xu, Touvron, Zarov, Ibarra, Kloumann, Misra,
  Evtimov, Zhang, Copet, Lee, Geffert, Vranes, Park, Mahadeokar, Shah, van~der
  Linde, Billock, Hong, Lee, Fu, Chi, Huang, Liu, Wang, Yu, Bitton, Spisak,
  Park, Rocca, Johnstun, Saxe, Jia, Alwala, Prasad, Upasani, Plawiak, Li,
  Heafield, Stone, El-Arini, Iyer, Malik, Chiu, Bhalla, Lakhotia,
  Rantala-Yeary, van~der Maaten, Chen, Tan, Jenkins, Martin, Madaan, Malo,
  Blecher, Landzaat, de~Oliveira, Muzzi, Pasupuleti, Singh, Paluri, Kardas,
  Tsimpoukelli, Oldham, Rita, Pavlova, Kambadur, Lewis, Si, Singh, Hassan,
  Goyal, Torabi, Bashlykov, Bogoychev, Chatterji, Zhang, Duchenne, Çelebi,
  Alrassy, Zhang, Li, Vasic, Weng, Bhargava, Dubal, Krishnan, Koura, Xu, He,
  Dong, Srinivasan, Ganapathy, Calderer, Cabral, Stojnic, Raileanu, Maheswari,
  Girdhar, Patel, Sauvestre, Polidoro, Sumbaly, Taylor, Silva, Hou, Wang,
  Hosseini, Chennabasappa, Singh, Bell, Kim, Edunov, Nie, Narang, Raparthy,
  Shen, Wan, Bhosale, Zhang, Vandenhende, Batra, Whitman, Sootla, Collot,
  Gururangan, Borodinsky, Herman, Fowler, Sheasha, Georgiou, Scialom,
  Speckbacher, Mihaylov, Xiao, Karn, Goswami, Gupta, Ramanathan, Kerkez,
  Gonguet, Do, Vogeti, Albiero, Petrovic, Chu, Xiong, Fu, Meers, Martinet,
  Wang, Wang, Tan, Xia, Xie, Jia, Wang, Goldschlag, Gaur, Babaei, Wen, Song,
  Zhang, Li, Mao, Coudert, Yan, Chen, Papakipos, Singh, Srivastava, Jain,
  Kelsey, Shajnfeld, Gangidi, Victoria, Goldstand, Menon, Sharma, Boesenberg,
  Baevski, Feinstein, Kallet, Sangani, Teo, Yunus, Lupu, Alvarado, Caples, Gu,
  Ho, Poulton, Ryan, Ramchandani, Dong, Franco, Goyal, Saraf, Chowdhury,
  Gabriel, Bharambe, Eisenman, Yazdan, James, Maurer, Leonhardi, Huang, Loyd,
  Paola, Paranjape, Liu, Wu, Ni, Hancock, Wasti, Spence, Stojkovic, Gamido,
  Montalvo, Parker, Burton, Mejia, Liu, Wang, Kim, Zhou, Hu, Chu, Cai, Tindal,
  Feichtenhofer, Gao, Civin, Beaty, Kreymer, Li, Adkins, Xu, Testuggine, David,
  Parikh, Liskovich, Foss, Wang, Le, Holland, Dowling, Jamil, Montgomery,
  Presani, Hahn, Wood, Le, Brinkman, Arcaute, Dunbar, Smothers, Sun, Kreuk,
  Tian, Kokkinos, Ozgenel, Caggioni, Kanayet, Seide, Florez, Schwarz, Badeer,
  Swee, Halpern, Herman, Sizov, {Guangyi}, {Zhang}, Lakshminarayanan, Inan,
  Shojanazeri, Zou, Wang, Zha, Habeeb, Rudolph, Suk, Aspegren, Goldman, Zhan,
  Damlaj, Molybog, Tufanov, Leontiadis, Veliche, Gat, Weissman, Geboski, Kohli,
  Lam, Asher, Gaya, Marcus, Tang, Chan, Zhen, Reizenstein, Teboul, Zhong, Jin,
  Yang, Cummings, Carvill, Shepard, McPhie, Torres, Ginsburg, Wang, Wu, U,
  Saxena, Khandelwal, Zand, Matosich, Veeraraghavan, Michelena, Li, Jagadeesh,
  Huang, Chawla, Huang, Chen, Garg, A, Silva, Bell, Zhang, Guo, Yu, Moshkovich,
  Wehrstedt, Khabsa, Avalani, Bhatt, Mankus, Hasson, Lennie, Reso, Groshev,
  Naumov, Lathi, Keneally, Liu, Seltzer, Valko, Restrepo, Patel, Vyatskov,
  Samvelyan, Clark, Macey, Wang, Hermoso, Metanat, Rastegari, Bansal,
  Santhanam, Parks, White, Bawa, Singhal, Egebo, Usunier, Mehta, Laptev, Dong,
  Cheng, Chernoguz, Hart, Salpekar, Kalinli, Kent, Parekh, Saab, Balaji,
  Rittner, Bontrager, Roux, Dollar, Zvyagina, Ratanchandani, Yuvraj, Liang,
  Alao, Rodriguez, Ayub, Murthy, Nayani, Mitra, Parthasarathy, Li, Hogan,
  Battey, Wang, Howes, Rinott, Mehta, Siby, Bondu, Datta, Chugh, Hunt, Dhillon,
  Sidorov, Pan, Mahajan, Verma, Yamamoto, Ramaswamy, Lindsay, Lindsay, Feng,
  Lin, Zha, Patil, Shankar, Zhang, Zhang, Wang, Agarwal, Sajuyigbe, Chintala,
  Max, Chen, Kehoe, Satterfield, Govindaprasad, Gupta, Deng, Cho, Virk,
  Subramanian, Choudhury, Goldman, Remez, Glaser, Best, Koehler, Robinson, Li,
  Zhang, Matthews, Chou, Shaked, Vontimitta, Ajayi, Montanez, Mohan, Kumar,
  Mangla, Ionescu, Poenaru, Mihailescu, Ivanov, Li, Wang, Jiang, Bouaziz,
  Constable, Tang, Wu, Wang, Wu, Gao, Kleinman, Chen, Hu, Jia, Qi, Li, Zhang,
  Zhang, Adi, Nam, {Yu}, {Wang}, Zhao, Hao, Qian, Li, He, Rait, DeVito,
  Rosnbrick, Wen, Yang, Zhao, and Ma}]{grattafioriLlama3Herd2024}
Aaron Grattafiori, Abhimanyu Dubey, Abhinav Jauhri, Abhinav Pandey, Abhishek
  Kadian, Ahmad Al-Dahle, Aiesha Letman, Akhil Mathur, Alan Schelten, Alex
  Vaughan, Amy Yang, Angela Fan, Anirudh Goyal, Anthony Hartshorn, Aobo Yang,
  Archi Mitra, Archie Sravankumar, Artem Korenev, Arthur Hinsvark, and 542
  others. 2024.
\newblock \href {https://arxiv.org/abs/2407.21783} {The llama 3 herd of
  models}.
\newblock ArXiv: 2407.21783 [cs.AI].

\bibitem[{Gugger et~al.(2022)Gugger, Debut, Wolf, Schmid, Mueller, Mangrulkar,
  Sun, and Bossan}]{guggerAccelerateTrainingInference2022}
Sylvain Gugger, Lysandre Debut, Thomas Wolf, Philipp Schmid, Zachary Mueller,
  Sourab Mangrulkar, Marc Sun, and Benjamin Bossan. 2022.
\newblock \href {https://github.com/huggingface/accelerate} {Accelerate:
  {Training} and inference at scale made simple, efficient and adaptable}.

\bibitem[{Guo et~al.(2025)Guo, Yang, Zhang, Song, Wang, Zhu, Xu, Zhang, Ma, Bi,
  Zhang, Yu, Wu, Wu, Gou, Shao, Li, Gao, Liu, Xue, Wang, Wu, Feng, Lu, Zhao,
  Deng, Ruan, Dai, Chen, Ji, Li, Lin, Dai, Luo, Hao, Chen, Li, Zhang, Xu, Ding,
  Gao, Qu, Li, Guo, Li, Chen, Yuan, Tu, Qiu, Li, Cai, Ni, Liang, Chen, Dong,
  Hu, You, Gao, Guan, Huang, Yu, Wang, Zhang, Zhao, Wang, Zhang, Xu, Xia,
  Zhang, Zhang, Tang, Zhou, Li, Wang, Li, Tian, Huang, Zhang, Wang, Chen, Du,
  Ge, Zhang, Pan, Wang, Chen, Jin, Chen, Lu, Zhou, Chen, Ye, Wang, Yu, Zhou,
  Pan, Li, Zhou, Wu, Yun, Pei, Sun, Wang, Zeng, Liu, Liang, Gao, Yu, Zhang,
  Xiao, An, Liu, Wang, Chen, Nie, Cheng, Liu, Xie, Liu, Yang, Li, Su, Lin, Li,
  Jin, Shen, Chen, Sun, Wang, Song, Zhou, Wang, Shan, Li, Wang, Wei, Zhang, Xu,
  Li, Zhao, Sun, Wang, Yu, Zhang, Shi, Xiong, He, Piao, Wang, Tan, Ma, Liu,
  Guo, Ou, Wang, Gong, Zou, He, Xiong, Luo, You, Liu, Zhou, Zhu, Huang, Li,
  Zheng, Zhu, Ma, Tang, Zha, Yan, Ren, Ren, Sha, Fu, Xu, Xie, Zhang, Hao, Ma,
  Yan, Wu, Gu, Zhu, Liu, Li, Xie, Song, Pan, Huang, Xu, Zhang, and
  Zhang}]{guoDeepSeekR1IncentivizesReasoning2025}
Daya Guo, Dejian Yang, Haowei Zhang, Junxiao Song, Peiyi Wang, Qihao Zhu,
  Runxin Xu, Ruoyu Zhang, Shirong Ma, Xiao Bi, Xiaokang Zhang, Xingkai Yu,
  Yu~Wu, Z.~F. Wu, Zhibin Gou, Zhihong Shao, Zhuoshu Li, Ziyi Gao, Aixin Liu,
  and 175 others. 2025.
\newblock \href {https://doi.org/10.1038/s41586-025-09422-z} {{DeepSeek}-{R1}
  incentivizes reasoning in {LLMs} through reinforcement learning}.
\newblock \emph{Nature}, 645(8081):633--638.

\bibitem[{Guo et~al.(2023)Guo, Guo, Nan, Liang, Guo, Chawla, Wiest, and
  Zhang}]{guoWhatCanLarge2023}
Taicheng Guo, kehan Guo, Bozhao Nan, Zhenwen Liang, Zhichun Guo, Nitesh Chawla,
  Olaf Wiest, and Xiangliang Zhang. 2023.
\newblock \href {https://doi.org/10.52202/075280-2607} {What can {Large}
  {Language} {Models} do in chemistry? {A} comprehensive benchmark on eight
  tasks}.
\newblock In \emph{Advances in neural information processing systems},
  volume~36, pages 59662--59688. Curran Associates, Inc.

\bibitem[{Hayes and Krippendorff(2007)}]{hayesAnsweringCallStandard2007}
Andrew~F. Hayes and Klaus Krippendorff. 2007.
\newblock \href {https://api.semanticscholar.org/CorpusID:15408575} {Answering
  the call for a standard reliability measure for coding data}.
\newblock \emph{Communication Methods and Measures}, 1:77 -- 89.

\bibitem[{Hsieh et~al.(2023)Hsieh, Li, Yeh, Nakhost, Fujii, Ratner, Krishna,
  Lee, and Pfister}]{hsiehDistillingStepbyStepOutperforming2023}
Cheng-Yu Hsieh, Chun-Liang Li, Chih-kuan Yeh, Hootan Nakhost, Yasuhisa Fujii,
  Alex Ratner, Ranjay Krishna, Chen-Yu Lee, and Tomas Pfister. 2023.
\newblock \href {https://doi.org/10.18653/v1/2023.findings-acl.507} {Distilling
  {Step}-by-{Step}! {Outperforming} {Larger} {Language} {Models} with {Less}
  {Training} {Data} and {Smaller} {Model} {Sizes}}.
\newblock In \emph{Findings of the {Association} for {Computational}
  {Linguistics}: {ACL} 2023}, pages 8003--8017, Toronto, Canada. Association
  for Computational Linguistics.

\bibitem[{Hu et~al.(2026)Hu, Zhu, Kellermann, Biddulph, Waiwitlikhit, Benn, and
  Kang}]{huBreakingBarriersReinforcement2026}
Chuxuan Hu, Yuxuan Zhu, Antony Kellermann, Caleb Biddulph, Suppakit
  Waiwitlikhit, Jason Benn, and Daniel Kang. 2026.
\newblock \href {https://openreview.net/forum?id=mvLhN0veUd} {Breaking
  {Barriers}: {Do} {Reinforcement} {Post} {Training} {Gains} {Transfer} {To}
  {Unseen} {Domains}?}
\newblock In \emph{The {Fourteenth} {International} {Conference} on {Learning}
  {Representations}}, Rio de Janeiro, Brazil.

\bibitem[{Hu et~al.(2022)Hu, Shen, Wallis, Allen-Zhu, Li, Wang, Wang, and
  Chen}]{huLoRALowrankAdaptation2022}
Edward~J. Hu, Yelong Shen, Phillip Wallis, Zeyuan Allen-Zhu, Yuanzhi Li, Shean
  Wang, Lu~Wang, and Weizhu Chen. 2022.
\newblock \href {https://openreview.net/forum?id=nZeVKeeFYf9} {{LoRA}:
  {Low}-rank adaptation of large language models}.
\newblock In \emph{International conference on learning representations}.

\bibitem[{Huan et~al.(2026)Huan, Li, Zheng, Xu, Kim, Du, Poovendran, Neubig,
  and Yue}]{huanDoesMathReasoning2026}
Maggie~Ziyu Huan, Yuetai Li, Tianyu Zheng, Xiaoyu Xu, Seungone Kim, Minxin Du,
  Radha Poovendran, Graham Neubig, and Xiang Yue. 2026.
\newblock \href {https://openreview.net/forum?id=sB6WgjbiJa} {Does {Math}
  {Reasoning} {Improve} {General} {LLM} {Capabilities}? {Understanding}
  {Transferability} of {LLM} {Reasoning}}.
\newblock In \emph{{ICLR} 2026 {Workshop} {ICBINB}}, Rio de Janeiro, Brazil.

\bibitem[{Ionov et~al.(2026)Ionov, Marshalova, and
  Malykh}]{ionovCallRewardRepeat2026}
Timur Ionov, Anna Marshalova, and Valentin Malykh. 2026.
\newblock \href {https://doi.org/10.18653/v1/2026.eacl-srw.21} {Call, {Reward},
  {Repeat}: {Advancing} {Dialog} {State} {Tracking} with {GRPO} and {Function}
  {Calling}}.
\newblock In \emph{Proceedings of the 19th {Conference} of the {European}
  {Chapter} of the {Association} for {Computational} {Linguistics} ({Volume} 4:
  {Student} {Research} {Workshop})}, pages 292--303, Rabat, Morocco.
  Association for Computational Linguistics.

\bibitem[{Islakoglu and Kalo(2025)}]{islakogluChronoSenseExploringTemporal2025}
Duygu~Sezen Islakoglu and Jan-Christoph Kalo. 2025.
\newblock \href {https://doi.org/10.18653/v1/2025.acl-short.46} {{ChronoSense}:
  {Exploring} {Temporal} {Understanding} in {Large} {Language} {Models} with
  {Time} {Intervals} of {Events}}.
\newblock In \emph{Proceedings of the 63rd {Annual} {Meeting} of the
  {Association} for {Computational} {Linguistics} ({Volume} 2: {Short}
  {Papers})}, pages 590--602, Vienna, Austria. Association for Computational
  Linguistics.

\bibitem[{James et~al.(1984)James, Demaree, and
  Wolf}]{jamesEstimatingWithingroupInterrater1984}
Lawrence~R. James, Robert~G. Demaree, and Gerrit Wolf. 1984.
\newblock \href {https://doi.org/10.1037/0021-9010.69.1.85} {Estimating
  within-group interrater reliability with and without response bias}.
\newblock \emph{Journal of Applied Psychology}, 69(1):85--98.

\bibitem[{Kwon et~al.(2023)Kwon, Li, Zhuang, Sheng, Zheng, Yu, Gonzalez, Zhang,
  and Stoica}]{kwonEfficientMemoryManagement2023}
Woosuk Kwon, Zhuohan Li, Siyuan Zhuang, Ying Sheng, Lianmin Zheng, Cody~Hao Yu,
  Joseph~E. Gonzalez, Hao Zhang, and Ion Stoica. 2023.
\newblock \href {https://doi.org/10.1145/3600006.3613165} {Efficient memory
  management for large language model serving with {PagedAttention}}.
\newblock In \emph{Proceedings of the 29th symposium on operating systems
  principles}, pages 611--626. Association for Computing Machinery.

\bibitem[{Li et~al.(2025)Li, Yue, Xu, Jiang, Niu, Lin, Ramasubramanian, and
  Poovendran}]{liSmallModelsStruggle2025}
Yuetai Li, Xiang Yue, Zhangchen Xu, Fengqing Jiang, Luyao Niu, Bill~Yuchen Lin,
  Bhaskar Ramasubramanian, and Radha Poovendran. 2025.
\newblock \href {https://doi.org/10.18653/v1/2025.findings-acl.1301} {Small
  {Models} {Struggle} to {Learn} from {Strong} {Reasoners}}.
\newblock In \emph{Findings of the {Association} for {Computational}
  {Linguistics}: {ACL} 2025}, pages 25366--25394, Vienna, Austria. Association
  for Computational Linguistics.

\bibitem[{Mangrulkar et~al.(2022)Mangrulkar, Gugger, Debut, Belkada, Paul, and
  Bossan}]{mangrulkarPEFTStateoftheartParameterefficient2022}
Sourab Mangrulkar, Sylvain Gugger, Lysandre Debut, Younes Belkada, Sayak Paul,
  and Benjamin Bossan. 2022.
\newblock \href {https://github.com/huggingface/peft} {{PEFT}:
  {State}-of-the-art parameter-efficient fine-tuning methods}.

\bibitem[{Mondorf and Plank(2024)}]{mondorfLiarLiarLogical2024}
Philipp Mondorf and Barbara Plank. 2024.
\newblock \href {https://doi.org/10.18653/v1/2024.emnlp-main.404} {Liar, liar,
  logical mire: a benchmark for suppositional reasoning in large language
  models}.
\newblock In \emph{Proceedings of the 2024 conference on empirical methods in
  natural language processing}, pages 7114--7137, Miami, Florida, USA.
  Association for Computational Linguistics.

\bibitem[{Qin et~al.(2021)Qin, Gupta, Upadhyay, He, Choi, and
  Faruqui}]{qinTIMEDIALTemporalCommonsense2021}
Lianhui Qin, Aditya Gupta, Shyam Upadhyay, Luheng He, Yejin Choi, and Manaal
  Faruqui. 2021.
\newblock \href {https://doi.org/10.18653/v1/2021.acl-long.549} {{TIMEDIAL}:
  {Temporal} {Commonsense} {Reasoning} in {Dialog}}.
\newblock In \emph{Proceedings of the 59th {Annual} {Meeting} of the
  {Association} for {Computational} {Linguistics} and the 11th {International}
  {Joint} {Conference} on {Natural} {Language} {Processing} ({Volume} 1: {Long}
  {Papers})}, pages 7066--7076, Online. Association for Computational
  Linguistics.

\bibitem[{Rajbhandari et~al.(2020)Rajbhandari, Rasley, Ruwase, and
  He}]{rajbhandariZeROMemoryOptimizations2020}
Samyam Rajbhandari, Jeff Rasley, Olatunji Ruwase, and Yuxiong He. 2020.
\newblock \href {https://doi.org/10.1109/SC41405.2020.00024} {{ZeRO}: {Memory}
  optimizations toward training trillion parameter models}.
\newblock In \emph{{SC20}: {International} conference for high performance
  computing, networking, storage and analysis}, pages 1--16. IEEE.

\bibitem[{Rameshkumar et~al.(2025)Rameshkumar, Huang, Sun, Xia, and
  Saparov}]{rameshkumarReasoningModelsReason2025}
Revanth Rameshkumar, Jimson Huang, Yunxin Sun, Fei Xia, and Abulhair Saparov.
  2025.
\newblock \href {https://doi.org/10.18653/v1/2025.ijcnlp-long.52} {Reasoning
  {Models} {Reason} {Well}, {Until} {They} {Don}'t}.
\newblock In \emph{Proceedings of the 14th {International} {Joint} {Conference}
  on {Natural} {Language} {Processing} and the 4th {Conference} of the
  {Asia}-{Pacific} {Chapter} of the {Association} for {Computational}
  {Linguistics}}, pages 936--956, Mumbai, India. The Asian Federation of
  Natural Language Processing and The Association for Computational
  Linguistics.

\bibitem[{Rasley et~al.(2020)Rasley, Rajbhandari, Ruwase, and
  He}]{rasleyDeepSpeedSystemOptimizations2020}
Jeff Rasley, Samyam Rajbhandari, Olatunji Ruwase, and Yuxiong He. 2020.
\newblock \href {https://doi.org/10.1145/3394486.3406703} {{DeepSpeed}:
  {System} optimizations enable training deep learning models with over 100
  billion parameters}.
\newblock In \emph{Proceedings of the 26th {ACM} {SIGKDD} international
  conference on knowledge discovery and data mining}, pages 3505--3506.
  Association for Computing Machinery.

\bibitem[{Shao et~al.(2024)Shao, Wang, Zhu, Xu, Song, Bi, Zhang, Zhang, Li, Wu,
  and Guo}]{shaoDeepSeekMathPushingLimits2024}
Zhihong Shao, Peiyi Wang, Qihao Zhu, Runxin Xu, Junxiao Song, Xiao Bi, Haowei
  Zhang, Mingchuan Zhang, Y.~K. Li, Y.~Wu, and Daya Guo. 2024.
\newblock \href {https://doi.org/10.48550/arXiv.2402.03300} {{DeepSeekMath}:
  {Pushing} the {Limits} of {Mathematical} {Reasoning} in {Open} {Language}
  {Models}}.
\newblock \emph{arXiv preprint}.
\newblock ArXiv:2402.03300 [cs].

\bibitem[{Skalse et~al.(2022)Skalse, Howe, Krasheninnikov, and
  Krueger}]{skalseDefiningCharacterizingReward2022}
Joar Skalse, Nikolaus Howe, Dmitrii Krasheninnikov, and David Krueger. 2022.
\newblock \href
  {https://proceedings.neurips.cc/paper_files/paper/2022/hash/3d719fee332caa23d5038b8a90e81796-Abstract-Conference.html}
  {Defining and {Characterizing} {Reward} {Gaming}}.
\newblock In \emph{Advances in {Neural} {Information} {Processing} {Systems}},
  volume~35, pages 9460--9471. Curran Associates, Inc.

\bibitem[{Su et~al.(2024)Su, Zhang, Zhu, Qu, Li, Zhang, and
  Cheng}]{suTimoBetterTemporal2024}
Zhaochen Su, Jun Zhang, Tong Zhu, Xiaoye Qu, Juntao Li, Min Zhang, and
  Yu~Cheng. 2024.
\newblock \href {https://openreview.net/forum?id=3TzGD95Jw1} {Timo: {Towards}
  {Better} {Temporal} {Reasoning} for {Language} {Models}}.
\newblock In \emph{First {Conference} on {Language} {Modeling}}.

\bibitem[{Tan et~al.(2023)Tan, Ng, and
  Bing}]{tanBenchmarkingImprovingTemporal2023}
Qingyu Tan, Hwee~Tou Ng, and Lidong Bing. 2023.
\newblock \href {https://doi.org/10.18653/v1/2023.acl-long.828} {Towards
  {Benchmarking} and {Improving} the {Temporal} {Reasoning} {Capability} of
  {Large} {Language} {Models}}.
\newblock In \emph{Proceedings of the 61st {Annual} {Meeting} of the
  {Association} for {Computational} {Linguistics} ({Volume} 1: {Long}
  {Papers})}, pages 14820--14835, Toronto, Canada. Association for
  Computational Linguistics.

\bibitem[{Tang et~al.(2025)Tang, Nuamah, and
  Belle}]{tangLTLBenchBenchmarksEvaluating2025}
Weizhi Tang, Kwabena Nuamah, and Vaishak Belle. 2025.
\newblock \href {https://doi.org/10.48550/arXiv.2407.05434} {{LTLBench}:
  {Towards} {Benchmarks} for {Evaluating} {Temporal} {Logic} {Reasoning} in
  {Large} {Language} {Models}}.
\newblock \emph{arXiv preprint}.
\newblock ArXiv:2407.05434 [cs.CL] version: 2.

\bibitem[{Tolan et~al.(2021)Tolan, Pesole, Martínez-Plumed,
  Fernández-Macías, Hernández-Orallo, and
  Gómez}]{tolanMeasuringOccupationalImpact2021}
Songül Tolan, Annarosa Pesole, Fernando Martínez-Plumed, Enrique
  Fernández-Macías, José Hernández-Orallo, and Emilia Gómez. 2021.
\newblock \href {https://doi.org/10.1613/jair.1.12647} {Measuring the
  {Occupational} {Impact} of {AI}: {Tasks}, {Cognitive} {Abilities} and {AI}
  {Benchmarks}}.
\newblock \emph{Journal of Artificial Intelligence Research}, 71:191--236.

\bibitem[{Varela et~al.(2026)Varela, Romero-Sorozabal, Rocon, and
  Cebrian}]{varelaRethinkingIllusionThinking2026}
Iñaki~Dellibarda Varela, Pablo Romero-Sorozabal, Eduardo Rocon, and Manuel
  Cebrian. 2026.
\newblock \href {https://doi.org/10.1007/978-3-032-11402-0_9} {Rethinking the
  illusion of thinking}.
\newblock In \emph{Artificial intelligence {XLII}}, volume 16301 of
  \emph{Lecture {Notes} in {Computer} {Science}}, pages 116--129, Cham.
  Springer Nature Switzerland.

\bibitem[{von Werra et~al.(2020)von Werra, Belkada, Tunstall, Beeching, Thrush,
  Lambert, Huang, Rasul, and
  Gallouédec}]{vonwerraTRLTransformerReinforcement2020}
Leandro von Werra, Younes Belkada, Lewis Tunstall, Edward Beeching, Tristan
  Thrush, Nathan Lambert, Shengyi Huang, Kashif Rasul, and Quentin Gallouédec.
  2020.
\newblock \href {https://github.com/huggingface/trl} {{TRL}: {Transformer}
  reinforcement learning}.

\bibitem[{Wang et~al.(2024)Wang, Ma, Zhang, Ni, Chandra, Guo, Ren, Arulraj, He,
  Jiang, Li, Ku, Wang, Zhuang, Fan, Yue, and Chen}]{wangMMLUproMoreRobust2024}
Yubo Wang, Xueguang Ma, Ge~Zhang, Yuansheng Ni, Abhranil Chandra, Shiguang Guo,
  Weiming Ren, Aaran Arulraj, Xuan He, Ziyan Jiang, Tianle Li, Max Ku, Kai
  Wang, Alex Zhuang, Rongqi Fan, Xiang Yue, and Wenhu Chen. 2024.
\newblock \href {https://doi.org/10.52202/079017-3018} {{MMLU}-pro: a more
  robust and challenging multi-task language understanding benchmark}.
\newblock In \emph{Advances in neural information processing systems},
  volume~37, pages 95266--95290. Curran Associates, Inc.

\bibitem[{Wang and Zhao(2024)}]{wangTRAMBenchmarkingTemporal2024}
Yuqing Wang and Yun Zhao. 2024.
\newblock \href {https://doi.org/10.18653/v1/2024.findings-acl.382} {{TRAM}:
  {Benchmarking} {Temporal} {Reasoning} for {Large} {Language} {Models}}.
\newblock In \emph{Findings of the {Association} for {Computational}
  {Linguistics}: {ACL} 2024}, pages 6389--6415, Bangkok, Thailand. Association
  for Computational Linguistics.

\bibitem[{Wei et~al.(2023)Wei, Su, Ma, Yu, Lei, Zhang, Zhao, and
  Liu}]{weiMenatQANewDataset2023}
Yifan Wei, Yisong Su, Huanhuan Ma, Xiaoyan Yu, Fangyu Lei, Yuanzhe Zhang, Jun
  Zhao, and Kang Liu. 2023.
\newblock \href {https://doi.org/10.18653/v1/2023.findings-emnlp.100}
  {{MenatQA}: {A} {New} {Dataset} for {Testing} the {Temporal} {Comprehension}
  and {Reasoning} {Abilities} of {Large} {Language} {Models}}.
\newblock In \emph{Findings of the {Association} for {Computational}
  {Linguistics}: {EMNLP} 2023}, pages 1434--1447, Singapore. Association for
  Computational Linguistics.

\bibitem[{White et~al.(2025)White, Dooley, Roberts, Pal, Feuer, Jain,
  Shwartz-Ziv, Jain, Saifullah, Dey, {Shubh-Agrawal}, Sandha, Naidu, Hegde,
  LeCun, Goldstein, Neiswanger, and
  Goldblum}]{whiteLiveBenchChallengingContaminationlimited2025}
Colin White, Samuel Dooley, Manley Roberts, Arka Pal, Benjamin Feuer,
  Siddhartha Jain, Ravid Shwartz-Ziv, Neel Jain, Khalid Saifullah, Sreemanti
  Dey, {Shubh-Agrawal}, Sandeep~Singh Sandha, Siddartha~Venkat Naidu, Chinmay
  Hegde, Yann LeCun, Tom Goldstein, Willie Neiswanger, and Micah Goldblum.
  2025.
\newblock \href {https://openreview.net/forum?id=sKYHBTAxVa} {{LiveBench}: a
  challenging, contamination-limited {LLM} benchmark}.
\newblock In \emph{The thirteenth international conference on learning
  representations}.

\bibitem[{Wolf et~al.(2020)Wolf, Debut, Sanh, Chaumond, Delangue, Moi, Cistac,
  Rault, Louf, Funtowicz, Davison, Shleifer, von Platen, Ma, Jernite, Plu, Xu,
  Le~Scao, Gugger, Drame, Lhoest, and
  Rush}]{wolfTransformersStateoftheartNatural2020}
Thomas Wolf, Lysandre Debut, Victor Sanh, Julien Chaumond, Clement Delangue,
  Anthony Moi, Pierric Cistac, Tim Rault, Rémi Louf, Morgan Funtowicz, Joe
  Davison, Sam Shleifer, Patrick von Platen, Clara Ma, Yacine Jernite, Julien
  Plu, Canwen Xu, Teven Le~Scao, Sylvain Gugger, and 3 others. 2020.
\newblock \href {https://doi.org/10.18653/v1/2020.emnlp-demos.6} {Transformers:
  {State}-of-the-art natural language processing}.
\newblock In \emph{Proceedings of the 2020 conference on empirical methods in
  natural language processing: {System} demonstrations}, pages 38--45.
  Association for Computational Linguistics.

\bibitem[{Xiong et~al.(2024)Xiong, Payani, Kompella, and
  Fekri}]{xiongLargeLanguageModels2024}
Siheng Xiong, Ali Payani, Ramana Kompella, and Faramarz Fekri. 2024.
\newblock \href {https://doi.org/10.18653/v1/2024.acl-long.563} {Large
  {Language} {Models} {Can} {Learn} {Temporal} {Reasoning}}.
\newblock In \emph{Proceedings of the 62nd {Annual} {Meeting} of the
  {Association} for {Computational} {Linguistics} ({Volume} 1: {Long}
  {Papers})}, pages 10452--10470, Bangkok, Thailand. Association for
  Computational Linguistics.

\bibitem[{Yang et~al.(2025)Yang, Li, Yang, Zhang, Hui, Zheng, Yu, Gao, Huang,
  Lv, Zheng, Liu, Zhou, Huang, Hu, Ge, Wei, Lin, Tang, Yang, Tu, Zhang, Yang,
  Yang, Zhou, Zhou, Lin, Dang, Bao, Yang, Yu, Deng, Li, Xue, Li, Zhang, Wang,
  Zhu, Men, Gao, Liu, Luo, Li, Tang, Yin, Ren, Wang, Zhang, Ren, Fan, Su,
  Zhang, Zhang, Wan, Liu, Wang, Cui, Zhang, Zhou, and
  Qiu}]{yangQwen3TechnicalReport2025}
An~Yang, Anfeng Li, Baosong Yang, Beichen Zhang, Binyuan Hui, Bo~Zheng, Bowen
  Yu, Chang Gao, Chengen Huang, Chenxu Lv, Chujie Zheng, Dayiheng Liu, Fan
  Zhou, Fei Huang, Feng Hu, Hao Ge, Haoran Wei, Huan Lin, Jialong Tang, and 41
  others. 2025.
\newblock \href {https://doi.org/10.48550/arXiv.2505.09388} {Qwen3 {Technical}
  {Report}}.
\newblock \emph{arXiv preprint}.
\newblock ArXiv:2505.09388 [cs].

\bibitem[{Zaradoukas et~al.(2026)Zaradoukas, Prenkaj, and
  Kasneci}]{zaradoukasReinforcementUnlearningGroup2026}
Efstratios Zaradoukas, Bardh Prenkaj, and Gjergji Kasneci. 2026.
\newblock \href {https://openreview.net/forum?id=BjWwqPE7mk} {Reinforcement
  unlearning via group relative policy optimization}.
\newblock In \emph{The fourteenth international conference on learning
  representations}.

\bibitem[{Zhong et~al.(2024)Zhong, Cui, Guo, Liang, Lu, Wang, Saied, Chen, and
  Duan}]{zhongAGIEvalHumancentricBenchmark2024}
Wanjun Zhong, Ruixiang Cui, Yiduo Guo, Yaobo Liang, Shuai Lu, Yanlin Wang, Amin
  Saied, Weizhu Chen, and Nan Duan. 2024.
\newblock \href {https://doi.org/10.18653/v1/2024.findings-naacl.149}
  {{AGIEval}: a human-centric benchmark for evaluating foundation models}.
\newblock In \emph{Findings of the association for computational linguistics:
  {NAACL} 2024}, pages 2299--2314, Mexico City, Mexico. Association for
  Computational Linguistics.

\bibitem[{Zhou et~al.(2026)Zhou, Pacchiardi, Martínez-Plumed, Collins,
  Moros-Daval, Zhang, Zhao, Huang, Sun, Prunty, Li, Sánchez-García,
  Jiang-Chen, Casares, Zu, Burden, Mehrbakhsh, Stillwell, Cebrian, Wang,
  Henderson, Wu, Kyllonen, Cheke, Xie, and
  Hernández-Orallo}]{zhouGeneralScalesUnlock2026}
Lexin Zhou, Lorenzo Pacchiardi, Fernando Martínez-Plumed, Katherine~M.
  Collins, Yael Moros-Daval, Seraphina Zhang, Qinlin Zhao, Yitian Huang, Luning
  Sun, Jonathan~E. Prunty, Zongqian Li, Pablo Sánchez-García, Kexin
  Jiang-Chen, Pablo A.~M. Casares, Jiyun Zu, John Burden, Behzad Mehrbakhsh,
  David Stillwell, Manuel Cebrian, and 7 others. 2026.
\newblock \href {https://doi.org/10.1038/s41586-026-10303-2} {General scales
  unlock {AI} evaluation with explanatory and predictive power}.
\newblock \emph{Nature}, 652(8108):58--67.

\end{thebibliography}

\clearpage
\appendix

\section{Demand level annotations (DeLeAn)}\label{app:delean}

Demand-Level Annotations (DeLeAn) are a set of general-scale, human-interpretable rubrics that map a benchmark item $X_j$ to a multi-dimensional \emph{demand profile} intended to capture which cognitive and knowledge demands a fully competent solver would need to meet in order to answer correctly~\citep{zhouGeneralScalesUnlock2026}. Concretely, each item is assigned an annotation vector $x^{(j)} \in \{0,\dots,5^+\}^{18}$, where each component $x_i^{(j)}$ is an open-scale demand level on dimension $i$ (0 = no meaningful demand; larger values indicate increasing demand; $5^+$ indicates demands beyond the highest anchor point of the rubric). In this paper, we use these demand vectors as structured, interpretable item-level features for analysing which cognitive capabilities current temporal reasoning benchmarks actually require.

The DeLeAn rubrics build upon seven broad capabilities proposed in \citep{tolanMeasuringOccupationalImpact2021}, which are refined into 11 core cognitive dimensions. In addition, the framework incorporates five knowledge-related dimensions and two extraneous ones (atypicality and volume) designed to capture task-specific complexity that goes beyond basic cognitive or knowledge demands. While the authors also define an additional extraneous dimension, unguessability, computed algorithmically based on prompt-specific characteristics, it is not considered in this work. Table \ref{tab:delean} summarises the dimensions included in the full rubric set.

All data used in this study is derived from \citealp{zhouGeneralScalesUnlock2026}. \Gls{tot}, and \gls{tcp} were labelled for this study and are licenced under CC BY 4.0 and MIT respectively.

\begin{table*}[!ht]
\centering
\scriptsize
\setlength{\tabcolsep}{4pt}
\renewcommand{\arraystretch}{0.95}
\resizebox{\textwidth}{!}{%
\begin{tabular}{@{}p{0.02\textwidth}p{0.2\textwidth} p{0.03\textwidth}p{0.20\textwidth} p{0.5\textwidth}@{}}
\toprule
\textbf{} & \textbf{Dimension (Broad)} & \textbf{} & \textbf{Dimension (Specific)} & \textbf{Demand description} \\
\midrule
\textbf{AS} & Attention and Scan & \textbf{AS} & Attention and Scan &
Focus on or locate specific elements within a given stream of information or environment in the whole process of solving a task. \\\midrule

\multirow{2}{*}{\textbf{CE}} & \multirow{2}{*}{Comprehension and}
& \textbf{CEc} & Verbal Comprehension &
Understand text, stories or the semantic content of other representations of ideas in different formats or modalities. \\ \cmidrule(lr){3-5}
& Expression & \textbf{CEe} & Verbal Expression &
Generate and articulate ideas, stories, or semantic content in different formats or modalities. \\\midrule

\textbf{CL} & Conceptualisation, Learning and Abstraction & \textbf{CL} & Conceptualisation, Learning and Abstraction &
Build new concepts, engage in inductive and analogical reasoning, map relationships between domains, and generate abstractions from concrete examples. \\\midrule

\multirow{3}{*}{MC} & \multirow{3}{*}{Metacognition and}
& \textbf{MCr} & Identifying Relevant Information &
Recognise what information helps solve the task or does not, and how this recognition process unfolds as they work toward the solution. \\\cmidrule(lr){3-5}
& Critical Thinking & \textbf{MCt} & Critical Thinking Processes &
Monitor or regulate multiple thought processes to answer the question effectively, ranging from simple recall to high-level critical thinking. \\\cmidrule(lr){3-5}
& & \textbf{MCu} & Calibrating Knowns and Unknowns &
Recognise the boundaries of one’s knowledge and confidently identify what one knows they know, knows they don’t know, or is uncertain about. \\\midrule

\textbf{MS} & Mind Modelling and Social Cognition & MS & Mind Modelling and Social Cognition &
Model the minds of other agents or reasoning about how the beliefs, desires, intentions, and emotions of multiple other agents might interact to determine future behaviours. \\\midrule

\multirow{2}{*}{\textbf{QL}} & \multirow{2}{*}{Quantitative and }
& \textbf{QLl} & Logical Reasoning &
Match and apply rules, procedures, algorithms or systematic steps to premises to solve problems, derive conclusions and make decisions. \\ \cmidrule(lr){3-5}
& Logical Reasoning & \textbf{QLq} & Quantitative Reasoning &
Work with and reason about quantities, numbers, and numerical relationships. \\\midrule

\textbf{SN} & Spatial Reasoning and Navigation & \textbf{SNs} & Spatio-physical Reasoning &
Understand spatial relationships between objects and predicting physical interactions. \\\midrule

\multirow{5}{*}{\textbf{KN}} & \multirow{5}{*}{Knowledge}
& \textbf{KNa} & Knowledge of Applied Sciences &
Knowledge or conceptual understanding in applied sciences (e.g., medicine, law, education, business, agriculture, engineering except IT). \\\cmidrule(lr){3-5}
& & \textbf{KNc} & Customary Everyday Knowledge &
Knowledge in information that most people in a given society typically acquire through daily life experiences, social interactions, and media. \\\cmidrule(lr){3-5}
& & \textbf{KNf} & Knowledge of Formal Sciences &
Knowledge or conceptual understanding in formal sciences (e.g., mathematics, logic, computer science, statistics). \\\cmidrule(lr){3-5}
& & \textbf{KNn} & Knowledge of Natural Sciences &
Knowledge or conceptual understanding in natural sciences (e.g., physics, chemistry, biology, astronomy, earth sciences, ecology). \\\cmidrule(lr){3-5}
& & \textbf{KNs} & Knowledge of Social Sciences &
Knowledge or conceptual understanding in social sciences and humanities (e.g., history, psychology, sociology, literature, art, philosophy). \\\midrule

\textbf{AT} & Atypicality & \textbf{AT} & Atypicality &
How uncommon the task is or how unlikely it is that the instance has appeared in various sources (internet, textbooks, tests). \\\midrule

\textbf{VO} & Volume & \textbf{VO} & Volume &
Proportional to the logarithm of the time a fully competent human needs to read and complete the task in ideal conditions, excluding interruptions. \\
\bottomrule
\end{tabular}
}
\caption{Overview of the DeLeAn demand dimensions, including broad cognitive and knowledge categories, specific sub-dimensions, and their semantic descriptions.}\label{tab:delean}
\end{table*}

\section{Data}

\subsection{Benchmark sizes}

Our experiments use two distinct data configurations. In the first configuration (see Section~\ref{sec:transfer}), we use existing benchmark labels to separate ToT and TRAM into semantic and arithmetic modes for the controlled transfer analysis, yielding four dataset-mode combinations. Their cardinalities are reported in Tab.~\ref{tab:base-dataset-cardinality}. In the second configuration (see Sections~\ref{sec:cognitive-demand} and~\ref{sec:drivers}), we annotate items from all nine datasets with \gls{delean} cognitive demand profiles, retaining TCP Long and Short as separate subsets. The resulting ten subsets are reported in Tab.~\ref{tab:delean-data-cardinality}. Because TRAM is substantially larger than the other benchmarks, we subsample it to ensure that no single benchmark dominates the training conditions. The cost of annotation with GPT-4o was also a factor in this decision.

\begin{table}
\centering
\begin{tabular}{lrrr}
\toprule
\textbf{Dataset} & \textbf{Train} & \textbf{Validation} & \textbf{Test} \\
\midrule
ToT Arithmetic & 1295 & 93 & 462 \\
ToT Semantic & 1959 & 141 & 700 \\
TRAM Arithmetic & 25499 & 2834 & 2700 \\
TRAM Semantic & 20121 & 2236 & 500 \\
\bottomrule
\end{tabular}
\caption{Train/validation/test cardinalities for the four dataset-mode combinations used in controlled transfer experiments, drawn from two benchmarks (ToT and TRAM).}\label{tab:base-dataset-cardinality}
\end{table}

\begin{table}
\centering
\begin{tabular}{lrrr}
\toprule
\textbf{Dataset} & \textbf{Train} & \textbf{Validation} & \textbf{Test} \\
\midrule
TRAM & 2736 & 322 & 739 \\
ToT Semantic & 2042 & 208 & 544 \\
ToT Arithmetic & 1297 & 141 & 401 \\
TimeQA & 530 & 58 & 139 \\
MenatQA & 490 & 49 & 141 \\
TempReason & 455 & 55 & 147 \\
Date Arithmetic & 347 & 41 & 105 \\
TimeDial & 256 & 30 & 54 \\
TCP Long & 207 & 21 & 57 \\
TCP Short & 170 & 23 & 43 \\
\bottomrule
\end{tabular}
\caption{DeLeAn-labelled dataset cardinalities (train/validation/test). TCP is split into Long and Short, yielding ten subsets from nine datasets.}
\label{tab:delean-data-cardinality}
\end{table}

\subsection{Partitioning TRAM into arithmetic and semantic tasks}\label{app:tram}

Unlike \gls{tot}, which provides an explicit arithmetic/semantic split, TRAM organises its tasks into categories such as foundational temporal understanding, temporal interpretation and computation, and advanced temporal concepts, rather than assigning a binary mode label. To derive an arithmetic/semantic partition for TRAM, we follow \citet{suTimoBetterTemporal2024}'s distinction between \emph{math-time tasks} and \emph{pure-time tasks}. TRAM's subtasks \emph{Frequency}, \emph{Duration}, \emph{Arithmetic}, and \emph{Ambiguity} are classified as math-time tasks, corresponding to the arithmetic mode. We therefore assign these four subtasks to the arithmetic split.

This leaves six remaining TRAM subtasks that can be classified as pure-time tasks, corresponding to the semantic mode. To keep the two splits balanced, however, we include only four subtasks in the semantic partition. TRAM's \emph{Ordering} and \emph{Storytelling} tasks align closely with the \gls{tot} Semantic subtasks \emph{FirstLast} and \emph{Timeline}, respectively, so we assign them to the semantic split.

This leaves \emph{Typical Time}, \emph{Causality}, \emph{Relation}, and \emph{NLI}. Because we include only two additional subtasks, we use \gls{qlq} and \gls{qll} scores as a tiebreaker. We select \emph{Relation} and \emph{NLI}, as these subtasks show the largest differences in \gls{qlq} and \gls{qll}. The scores are reported in Tab.~\ref{tab:tram-qlq-qll}.

\begin{table*}
\centering
\begin{tabular}{llrrr}
\toprule
\textbf{Coarse task} & \textbf{Task} & \textbf{QLq} & \textbf{QLl} & \textbf{$\text{QLq} - \text{QLl}$} \\
\midrule
Arithmetic & Arithmetic & 1.40 & 1.01 & 0.39 \\
Arithmetic & Frequency & 0.77 & 0.55 & 0.22 \\
Arithmetic & Duration & 0.86 & 0.66 & 0.20 \\
Arithmetic & Ambiguity Resolution & 1.19 & 1.07 & 0.12 \\
--- & Typical Time & 0.10 & 0.25 & -0.15 \\
Semantic & Ordering & 0.20 & 0.86 & -0.66 \\
Semantic & Storytelling & 0.02 & 0.95 & -0.93 \\
--- & Causality & 0.06 & 1.00 & -0.94 \\
Semantic & Relation & 0.02 & 1.01 & -0.99 \\
Semantic & NLI & 0.12 & 1.13 & -1.01 \\
\bottomrule
\end{tabular}
\caption{Average \gls{qlq} and \gls{qll} scores and their difference for TRAM tasks.}
\label{tab:tram-qlq-qll}
\end{table*}

\begin{figure}
\centering
\import{figures/}{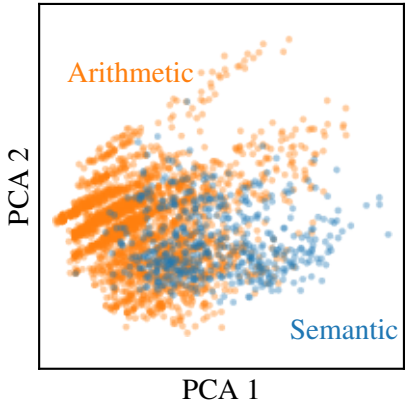}
\caption{\gls{pca} of TRAM items coloured by their assigned mode. The two groups show substantial overlap, confirming that the arithmetic/semantic partition is not cleanly reflected in TRAM's cognitive demand structure.}\label{fig:tram_pca}
\end{figure}

To further assess the quality of this partition, we project all TRAM items' demand profiles into a two-dimensional \gls{pca} space and colour them by assigned mode (Fig.~\ref{fig:tram_pca}). The resulting separation is weak, consistent with the findings in Section~\ref{sec:cognitive-demand} that arithmetic and semantic tasks are not strongly distinguishable by demand vectors.

\subsection{Cognitive demand profiles of ToT Semantic question types}\label{app:tot-semantic}

The ToT Semantic subset contains eight question types that vary in their reliance on quantitative reasoning. Tab.~\ref{tab:tot-semantic-question-types} reports the average \gls{qlq} and \gls{qll} scores for each type. Two question types, \emph{RelationDuration} and \emph{Timeline}, have a positive \gls{qlq}$-$\gls{qll} difference, indicating that they place slightly higher quantitative than logical demands despite being labelled as semantic. Conversely, \emph{FirstLast}, \emph{BeforeAfter}, and \emph{EventAtTimeOfOtherEvent} show the largest negative differences, consistent with their reliance on relational reasoning. This within-subset heterogeneity supports our finding (Section~\ref{sec:cognitive-demand}) that the semantic label subsumes tasks with meaningfully different cognitive profiles.

\begin{table*}
\centering
\begin{tabular}{lrrr}
\toprule
\textbf{Question types} & \textbf{QLq} & \textbf{QLl} & \textbf{$\text{QLq} - \text{QLl}$} \\
\midrule
RelationDuration & 2.04 & 1.90 & 0.13 \\
Timeline & 2.15 & 2.07 & 0.08 \\
EventAtTimeT & 1.76 & 1.94 & -0.18 \\
EventAtWhatTime & 1.22 & 1.43 & -0.21 \\
NumberOfEventsInTimeInterval & 2.58 & 2.85 & -0.27 \\
EventAtTimeOfOtherEvent & 2.39 & 2.84 & -0.45 \\
BeforeAfter & 2.45 & 2.90 & -0.45 \\
FirstLast & 2.05 & 2.57 & -0.52 \\
\bottomrule
\end{tabular}
\caption{Average \gls{qlq} and \gls{qll} scores and their difference for \gls{tot} Semantic question types}\label{tab:tot-semantic-question-types}
\end{table*}

\subsection{Per-benchmark cognitive demand profiles}\label{app:benchmark-profiles}

\begin{figure*}[!ht]
\centering
\import{figures/}{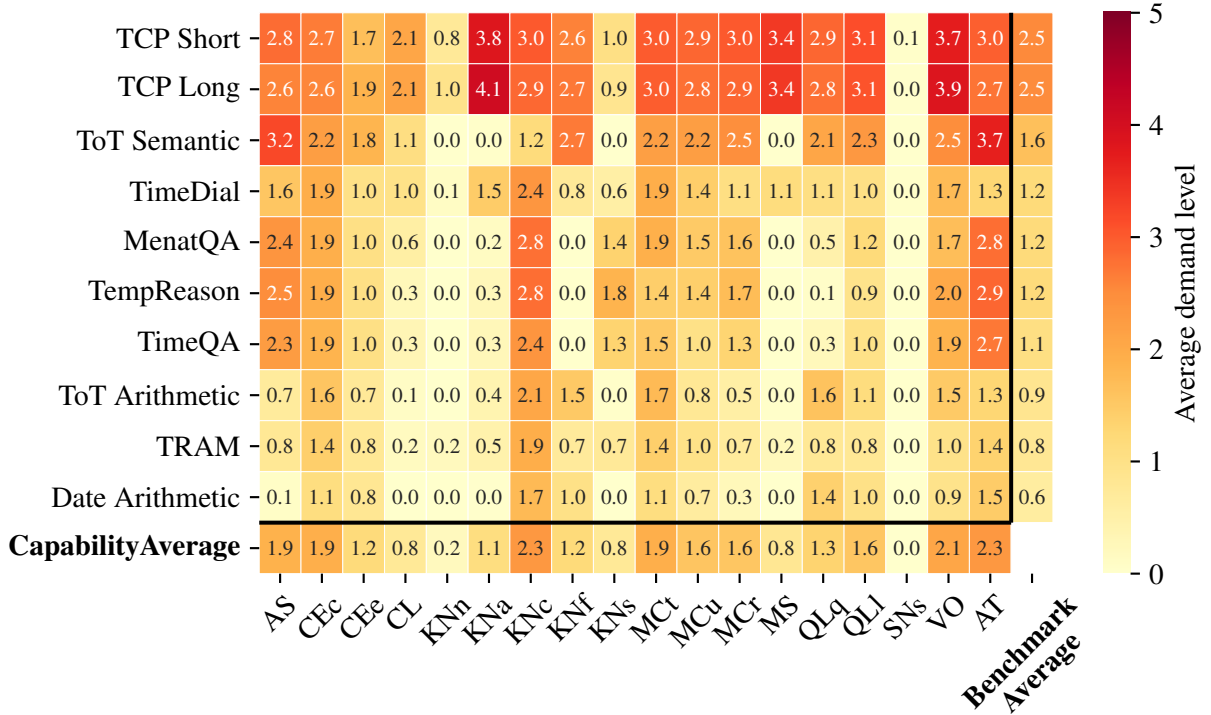}
\caption{Average demand level per cognitive ability for the ten subsets of the nine datasets, with TCP Long and Short shown separately. Subsets are sorted by overall average demand level.}\label{fig:benchmark_heatmap}
\end{figure*}

Fig.~\ref{fig:benchmark_heatmap} shows the average demand level per cognitive dimension for each of the ten dataset subsets, sorted by overall difficulty (computed as the mean across all 18 dimensions). The most demanding subsets (TCP Short and TCP Long) show elevated scores across nearly all dimensions, while the least demanding (Date Arithmetic) has near-zero demand on most dimensions except \gls{qlq} and \gls{knc}. Notably, \gls{qlq} and \gls{qll} are not among the highest-demand dimensions for any subset. Instead, \gls{knc}, and \gls{cec}, among others, dominate. This pattern is consistent with our main finding that temporal benchmarks are knowledge-heavy rather than reasoning-heavy.

\begin{figure*}
\centering
\includegraphics[width=0.9\linewidth]{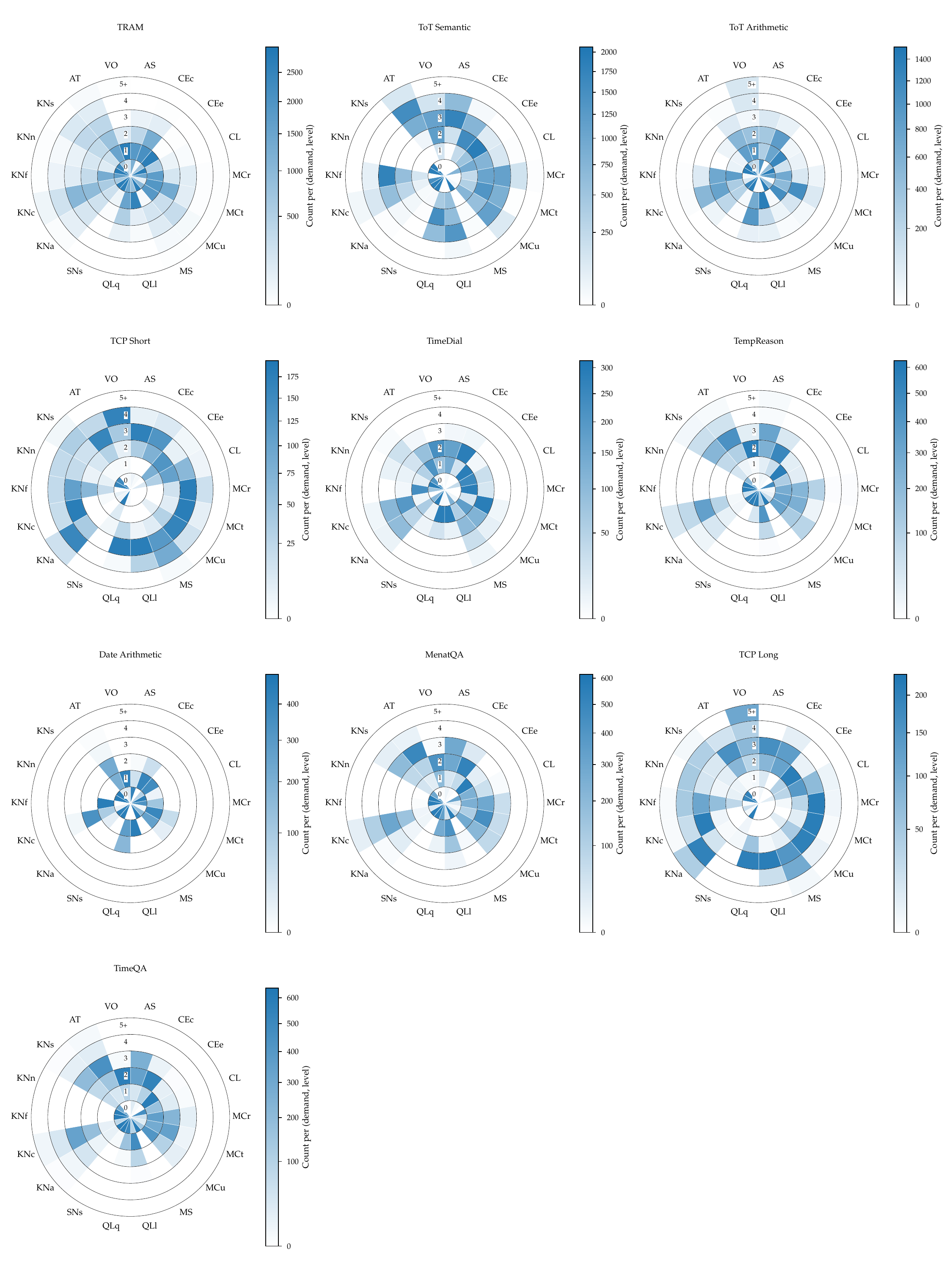}
\caption{Demand level counts per cognitive ability for the ten dataset subsets, with TCP Long and Short shown separately.}\label{fig:benchmark_pies}
\end{figure*}

Fig.~\ref{fig:benchmark_pies} provides a complementary view, showing the full distribution of demand levels (0 through 5+) for each cognitive dimension within each of the ten dataset subsets. These distributions confirm that most subsets concentrate their \gls{qlq} and \gls{qll} demands at levels 0--2, with few items reaching the higher levels that would be expected if genuine arithmetic or logical reasoning were the primary bottleneck.

\section{Metrics}\label{app:metrics}

To reduce spurious mismatches caused by formatting variability, we evaluate model outputs with benchmark-specific post-processing before computing final scores. Across our evaluation modules, normalisation generally takes one of three forms: extracting the answer from a larger generated response, canonicalising the representation of the answer, and then comparing either normalised strings, normalised structured objects, or normalised sets. The exact procedure differs by benchmark.

\subsection{TRAM accuracy}

For TRAM, we normalise predictions by extracting the final option letter from responses matching the template ``The final answer is (X)'', with \(\text{X} \in \{A,B,C,D\}\). We then compute exact-match accuracy against the gold option label. This removes variability in preceding reasoning text but still requires a standardised final-answer template.

\subsection{TCP accuracy}

For TCP, the evaluator expects the model's final answer to appear in boxed LaTeX notation, using the pattern \verb|\boxed{...}|. It extracts the boxed span with the regular expression \verb|\\boxed\{([^}]*)\}| and strips surrounding white space from the extracted content. Comparison is then performed by exact string match.

In the TCP Short subset only, the evaluator removes the literal string ``GMT'' from both the prediction and the reference before comparison; no analogous normalisation is applied for TCP Long. The system prompt includes the exact expected answer format.

\subsection{Test of Time}

For \gls{tot}, we extract the first valid JSON object from the model output, after escaping unescaped control characters inside JSON strings. We discard the explanation field and compare only the answer-bearing content. In the Arithmetic subset, we additionally normalise schema mismatches such as age versus answer, lowercase items in \texttt{ordered\_list} and \texttt{unordered\_list} answer types, cast prediction values to the reference types when possible, and sort \texttt{unordered\_list} values to make comparison order-invariant. In the Semantic subset, we normalise both prediction and reference by converting them to strings before exact comparison. 

\subsection{TimeBench derived benchmarks}

For TimeBench tasks, we first extract the final answer from the text following the marker ``Thus, the correct answer is:''. We retain only the first subsequent line, strip trailing sentence punctuation, and canonicalise any response containing unanswerable to the label unanswerable. For TempReason, TimeQA, and MenatQA, the extracted answer is scored with standard SQuAD exact match and F1. For Date Arithmetic, predictions and references are parsed into date objects and normalised by setting the day to 1, so comparison is effectively at the month-year level. For TimeDial, prediction and reference are mapped to sets of selected options in {A,B,C,D} and scored using set exact match and set-based F1.

\subsection{Non-temporal multiple-choice accuracy}

For single-answer multiple-choice examples from the Civil Service Examination, LSAT, MMLU-Pro, and the relevant Math tasks, we inspect the final 300 characters of the completion. We preferentially extract the final option letter in \(\{\mathrm{A},\ldots,\mathrm{J}\}\) appearing in an answer-bearing expression such as ``the answer is X'', ``answer: X'', ``option X'', or ``choice X'', or as a standalone line. If no such expression is found, we fall back to the final isolated letter in this range. The extracted letter is uppercased and compared by exact match with the gold option label.

\subsection{TruthQuest accuracy}

For TruthQuest, the reference annotation is converted to the set of option labels corresponding to knights. From the final 300 characters of the completion, we collect option letters in answer-bearing expressions; if none are present, we collect all isolated letters in \(\{\mathrm{A},\ldots,\mathrm{J}\}\). Both prediction and reference are treated as sets, so ordering and duplicate labels are ignored. A reward of 1 is assigned only when the predicted and reference sets are identical.

\subsection{Math and OmniMath accuracy}

Math examples use several answer representations. Multiple-choice items follow the option-letter procedure above. For integer-valued Math and OmniMath items, we extract the final integer appearing in the last 200 characters of the completion, cast it to an integer, and compare it exactly with the reference. For sequence-valued Math items, we extract the last valid JSON array from the complete response and compare it directly with the reference list; this comparison is order-sensitive.

For fraction-valued OmniMath items, we parse a LaTeX fraction written using \verb|\frac|, \verb|\dfrac|, or \verb|\tfrac|. If no valid LaTeX fraction is found, we use the final plain-text fraction of the form \(p/q\). The numerator and denominator are converted to an exact rational number and reduced to the canonical string \(p/q\) before exact comparison with the similarly normalised reference.

\subsection{ChemLLMBench and Reasoning accuracy}

For the SMILES-valued ChemLLMBench examples, we scan the completion from right to left for a whitespace-delimited SMILES-like token that can be parsed by RDKit. The predicted molecule is converted to canonical isomeric SMILES and compared by exact string match with the canonicalised reference.

For free-text Reasoning examples, we inspect the final 300 characters of the completion, lowercase the text, remove punctuation, and collapse consecutive whitespace. We apply the same normalisation to the reference and award a match when the normalised reference occurs as a substring of the normalised completion.

\section{Training settings}
\label{app:training-settings}

We used Transformers 4.57.3 \citep{wolfTransformersStateoftheartNatural2020} to load the models, TRL 0.27.2 \citep{vonwerraTRLTransformerReinforcement2020} for \gls{grpo}, and PEFT 0.18.0 \citep{mangrulkarPEFTStateoftheartParameterefficient2022} for LoRA fine-tuning \citep{huLoRALowrankAdaptation2022}. LoRA used rank \(r=16\) and scaling parameter \(\alpha=32\) and was applied to the \texttt{q\_proj}, \texttt{k\_proj}, \texttt{v\_proj}, and \texttt{o\_proj} modules. The maximum \gls{grpo} completion length was 3,000 tokens.

All GRPO runs used a learning rate of \(5\times10^{-5}\) with linear decay, training seed 42, no KL penalty (\(\beta=0\)), and no warmup. Dataset splits used seed 6; the temporal mixture's initial stratified train/test split used seed 42.

Training used PyTorch 2.9.0 \citep{anselPyTorch2Faster2024}, Accelerate 1.12.0 \citep{guggerAccelerateTrainingInference2022}, and DeepSpeed 0.18.3 \citep{rasleyDeepSpeedSystemOptimizations2020}. Regular single-node training used DeepSpeed ZeRO-2 , whereas Qwen3-32B training used ZeRO-3 \citep{rajbhandariZeROMemoryOptimizations2020}. Evaluation was performed using vLLM 0.11.2 \citep{kwonEfficientMemoryManagement2023}.

DeLeAn-filtered and non-temporal training runs allocated one GPU for training and one GPU for vLLM generation. Single-dataset training runs allocated two GPUs for training and one GPU for generation. Qwen3-32B training used three compute nodes equipped with four NVIDIA GH200 GPUs each: eight GPUs across two nodes were used for training, while four GPUs on the third node were used for generation. Evaluation used one GPU per run. Full training runs generally took several hours, with longer runs approaching or exceeding 12 hours. Individual evaluations generally took approximately 1--8 minutes, although some larger evaluations took longer.

\section{Label reliability}\label{app:label reliability}

We test label agreement using the within-group interrater reliability, which is the main metric used for the \gls{delean} paper, between the labels from GPT-4o and Gemini 3.1 Flash Lite. Even though this is the metric reported in the original paper ~\citep{zhouGeneralScalesUnlock2026}, we additionally report exact score matches, within one matches, and Krippendorff's alpha.

For each item, within-group interrater agreement is
\[
r_{WG}=1-\frac{s_{\mathrm{obs}}^2}{s_E^2}, \qquad
s_{\mathrm{obs}}^2=\frac{(x_{\mathrm{gemini}}-x_{\mathrm{gpt}})^2}{4}.
\]
The reference variance is \(s_E^2=(A^2-1)/12=35/12\), assuming a discrete uniform distribution over \(A=6\) equally spaced category codes \(\{0,1,2,3,4,5\}\). The open-ended top category \(5^+\) is encoded as 5. We use all six rubric categories for this reference distribution, even when fewer levels are observed.

We use the population variance across the two annotators, clip each item's \(r_{WG}\) to \([0,1]\), and report the mean over items with both labels present in each dimension.

The capability analyses retain dimensions with at least three observed demand levels and at least 30 items in the dominant slices (Section~\ref{sec:evaluation}). The focus of our work lies in inspecting \gls{qll} and \gls{qlq}. However, we are also interested in confirming whether the improvements observed for \gls{cec}, \gls{knc}, and \gls{kns} in Fig.~\ref{fig:heatmap_dw_oma_adele} are not due to label biases but can be reproduced by Gemini 3.1 Flash Lite.

Tab.~\ref{tab:gemini-gpt4o-agreement} shows that overall agreement according to \(r_{WG}\) is high with values close to the maximum of 1. \Gls{knc} is an outlier, showing a lower agreement of only 0.66. \Gls{cec}, while scoring high for \(r_{WG}\), has a lower value for Krippendorff's alpha. The issue is that both models assign items a demand level of 1--2. Absolute closeness is therefore high, explaining the high value for \(r_{WG}\), but on the other hand, it increases the by-chance agreement, which Krippendorff's alpha penalises.

\begin{table*}
\centering
\begin{tabular}{lrrrr}
\toprule
\textbf{Dimension} & \textbf{Exact (\%)} & \textbf{Within one (\%)} & \textbf{Ordinal $\alpha$} & \textbf{$r_{\mathrm{WG}}$} \\
\midrule
KNc & 27 & 45 & -0.23 & 0.66 \\
CEe & 62 & 98 & -0.01 & 0.96 \\
SNs & 99 & 100 & 0.06 & 1.00 \\
MCu & 42 & 85 & 0.34 & 0.90 \\
CL & 70 & 93 & 0.42 & 0.95 \\
CEc & 60 & 99 & 0.44 & 0.96 \\
KNa & 83 & 92 & 0.44 & 0.94 \\
KNf & 62 & 78 & 0.47 & 0.87 \\
KNn & 96 & 98 & 0.53 & 0.99 \\
MCt & 65 & 99 & 0.56 & 0.97 \\
KNs & 79 & 89 & 0.57 & 0.93 \\
AT & 37 & 81 & 0.59 & 0.88 \\
VO & 60 & 97 & 0.60 & 0.96 \\
QLl & 68 & 95 & 0.68 & 0.96 \\
MS & 92 & 95 & 0.69 & 0.97 \\
MCr & 56 & 94 & 0.71 & 0.94 \\
QLq & 63 & 97 & 0.74 & 0.96 \\
AS & 46 & 94 & 0.76 & 0.94 \\
\bottomrule
\end{tabular}
\caption{Inter-rater agreement between Gemini 3.1 Flash Lite and GPT-4o demand labels.}
\label{tab:gemini-gpt4o-agreement}
\end{table*}

Tab.~\ref{tab:gemini-gpt4o-demand-levels} compares the mean demand levels between Gemini and GPT. We can see that the low agreement in \gls{knc} can be explained by GPT-4o assigning much higher demands (1.59 higher in average).

\begin{table*}
\centering
\begin{tabular}{lrrr}
\toprule
\textbf{Dimension} & \textbf{Gemini} & \textbf{GPT-4o} & \textbf{GPT-4o $-$ Gemini} \\
\midrule
KNc & 0.36 & 1.96 & 1.59 \\
KNf & 0.61 & 1.27 & 0.66 \\
MCu & 0.73 & 1.36 & 0.63 \\
AT & 1.67 & 2.19 & 0.53 \\
AS & 1.16 & 1.67 & 0.52 \\
KNs & 0.26 & 0.57 & 0.30 \\
CL & 0.23 & 0.52 & 0.29 \\
KNa & 0.19 & 0.47 & 0.28 \\
VO & 1.46 & 1.70 & 0.24 \\
MCt & 1.58 & 1.75 & 0.17 \\
QLl & 1.21 & 1.37 & 0.15 \\
MS & 0.16 & 0.25 & 0.09 \\
QLq & 1.17 & 1.25 & 0.08 \\
KNn & 0.04 & 0.10 & 0.06 \\
MCr & 1.31 & 1.33 & 0.03 \\
CEe & 1.11 & 1.11 & 0.01 \\
SNs & 0.01 & 0.01 & 0.01 \\
CEc & 1.77 & 1.76 & -0.00 \\
\bottomrule
\end{tabular}
\caption{Mean demand level per dimension for Gemini 3.1 Flash Lite and GPT-4o annotations.}
\label{tab:gemini-gpt4o-demand-levels}
\end{table*}

Gemini treats \gls{knc} as the everyday knowledge a solver needs to bring to the task. When the prompt supplies the information required to obtain the answer, Gemini treats it as requiring little or no prior everyday knowledge.

GPT-4o appears to score the topical complexity of the prompt (e.g., German political parties, medieval history, museum associations), conflating the necessity of specialised content with the requirement of everyday knowledge---which the rubric explicitly excludes (specialised academic knowledge is not \gls{knc}).

The main findings still reproduce with Gemini 3.1 Flash Lite, even with its lower \gls{knc} scores. Instead of \gls{knc}, \gls{kns} and in one instance \gls{knf} show the largest increase in capability after finetuning across all training settings. Most importantly, \gls{qll} and \gls{qlq} barely improve (Fig.~\ref{fig:heatmap_dw_oma_adele_gemini}). The same pattern also holds in the one-minus-accuracy setting, except for \gls{at}, which shows the strongest decrease in the $\mathrm{QLl}$ = $\mathrm{QLq}+2$ setting. \Gls{cec} does not survive this ablation. However, although, Tab.~\ref{tab:gemini-gpt4o-demand-levels} shows strong agreement in mean demand for \gls{cec}, annotation agreement shows lower values (Tab.~\ref{tab:gemini-gpt4o-agreement}). The reason is that GPT-4o assigns demand 3 levels more often than Gemini 3.1 Flash Lite. As a result, the potential gains in capability are capped more often at level 2.

\begin{figure*}[t]
    \centering
    (a) Reward: Accuracy\par
    \import{figures/}{heatmap_dw_adele_gemini.pgf}

    \vspace{0pt}
    \noindent\rule{0.9\linewidth}{0.1pt}
    \vspace{0pt}

    (b) Reward: One-minus-accuracy\par
    \import{figures/}{heatmap_oma_adele_gemini.pgf}

    \caption{Average absolute change in AUC between the untrained model (baseline) and the trained model evaluated using Gemini 3.1 Flash Lite \gls{delean} labels. Results are averaged over six models from three model families and two sizes. Panel (a) shows the accuracy reward setting, and panel (b) shows the one-minus-accuracy reward setting.}
    \label{fig:heatmap_dw_oma_adele_gemini}
\end{figure*}

\section{Per-model results}\label{app:per-model}

The transfer results reported in Section~\ref{sec:transfer} are averaged across six models. Tab.~\ref{tab:model-performance-deltas} disaggregates these results, reporting the baseline accuracy range (min, median, max across models), the mean and median performance delta, and the number of models that improved versus worsened for each train--test combination under both reward settings.

Several patterns are worth noting. First, in-domain gains under the accuracy reward are consistent across all six models: every model improves on its own training set (6/0 in all diagonal cells). Second, the absence of cross-benchmark transfer is not driven by individual outliers. For most off-diagonal cells, the number of models that improve versus worsen is close to 3/3, indicating that the lack of transfer is a stable property of the task structure rather than an artefact of averaging. Third, under the one-minus-accuracy reward, unlearning is similarly concentrated: for example, training on \gls{tot} Arithmetic degrades all six models on \gls{tot} Arithmetic (0/6) and five of six on TRAM Arithmetic (1/5), confirming that the unlearning signal is reliably absorbed. Finally, the wide baseline range (e.g., 25.74--87.74 for TRAM Arithmetic) reflects genuine variation in model strength, yet the direction of transfer effects is consistent across this range, further supporting the robustness of our conclusions.

This appendix also collects the per-model results for the item-level experiments of Section~\ref{sec:drivers}, which Fig.~\ref{fig:heatmap_dw_oma_adele} reports only as averages. Tables~\ref{tab:gemma-3-4b-dw-ability-auc}--\ref{tab:qwen3-8b-oma-ability-auc} report, for each of the six models, the baseline capability \gls{auc} and its change under each of the six training conditions, separately for the accuracy and one-minus-accuracy reward settings.

\begin{table*}
\centering
\begin{tabular}{lrrrrrrr}
\toprule
 &  & \multicolumn{6}{c}{\textbf{Capability AUC $\Delta$ vs.\ Baseline}} \\
\cmidrule(lr){3-8}
 & \multicolumn{1}{l}{\textbf{Baseline}} & \textbf{QLl$=0$} & \textbf{QLq$=0$} & \multicolumn{2}{c}{\textbf{QLl$>$QLq}} & \multicolumn{2}{c}{\textbf{QLq$>$QLl}} \\
\cmidrule(lr){5-8}
 &  &  &  & \textbf{+1} & \textbf{+2} & \textbf{+1} & \textbf{+2} \\
\midrule
AS & 2.10 & 0.05 & 0.25 & 0.21 & 0.28 & 0.39 & 0.11 \\
AT & 2.69 & 0.10 & 0.16 & 0.17 & 0.20 & 0.05 & 0.01 \\
CEc & 2.93 & -0.30 & 0.41 & -0.08 & 0.16 & -0.34 & -0.09 \\
KNa & 3.08 & 0.05 & 0.05 & -0.07 & -0.04 & 0.08 & -0.12 \\
KNc & 2.84 & 0.04 & 0.36 & 0.04 & 0.17 & 0.22 & 0.05 \\
KNf & 2.59 & 0.01 & 0.05 & 0.22 & 0.35 & 0.11 & 0.34 \\
KNs & 2.83 & 0.20 & 0.09 & 0.07 & -0.02 & 0.04 & 0.03 \\
MCr & 1.93 & 0.01 & 0.33 & 0.16 & 0.17 & 0.51 & 0.04 \\
MCt & 2.61 & -0.21 & 0.37 & -0.03 & 0.04 & -0.05 & -0.07 \\
MCu & 2.30 & 0.25 & 0.37 & -0.29 & 0.29 & 0.24 & -0.02 \\
MS & \textbf{3.20} & -0.13 & -0.11 & -0.07 & -0.05 & 0.02 & 0.02 \\
QLl & 2.07 & \textbf{0.38} & \textbf{0.62} & \textbf{0.72} & \textbf{0.80} & 0.61 & \textbf{0.60} \\
QLq & 2.91 & -0.26 & 0.07 & -0.25 & -0.21 & -0.31 & -0.12 \\
VO & 1.83 & 0.10 & 0.21 & 0.21 & 0.13 & \textbf{0.64} & 0.34 \\
\bottomrule
\end{tabular}
\caption{Capability \gls{auc} change ($\Delta$) for gemma-3-4b in the accuracy reward setting. Highest values per condition are bold.}
\label{tab:gemma-3-4b-dw-ability-auc}
\end{table*}

\begin{table*}
\centering
\begin{tabular}{lrrrrrrr}
\toprule
 &  & \multicolumn{6}{c}{\textbf{Capability AUC $\Delta$ vs.\ Baseline}} \\
\cmidrule(lr){3-8}
 & \multicolumn{1}{l}{\textbf{Baseline}} & \textbf{QLl$=0$} & \textbf{QLq$=0$} & \multicolumn{2}{c}{\textbf{QLl$>$QLq}} & \multicolumn{2}{c}{\textbf{QLq$>$QLl}} \\
\cmidrule(lr){5-8}
 &  &  &  & \textbf{+1} & \textbf{+2} & \textbf{+1} & \textbf{+2} \\
\midrule
AS & 2.10 & -0.16 & -0.21 & 0.30 & 0.01 & -0.47 & -0.01 \\
AT & 2.69 & -0.47 & \textbf{-0.44} & 0.03 & 0.00 & -0.45 & -0.02 \\
CEc & 2.93 & -1.28 & -0.27 & -0.31 & 0.09 & -0.64 & -0.22 \\
KNa & 3.08 & -1.05 & -0.14 & 0.17 & 0.01 & -0.24 & -0.22 \\
KNc & 2.84 & -1.08 & -0.13 & 0.14 & 0.07 & -0.19 & 0.08 \\
KNf & 2.59 & -0.46 & 0.11 & -0.04 & 0.54 & -0.82 & -0.06 \\
KNs & 2.83 & \textbf{-1.51} & -0.25 & 0.26 & 0.27 & 0.00 & 0.10 \\
MCr & 1.93 & -0.45 & 0.16 & 0.47 & 0.16 & 0.11 & 0.19 \\
MCt & 2.61 & -1.02 & -0.07 & 0.09 & 0.34 & -0.73 & -0.10 \\
MCu & 2.30 & -0.83 & -0.03 & 0.25 & 0.36 & -0.20 & -0.24 \\
MS & 3.20 & -0.66 & -0.15 & -0.21 & \textbf{-0.11} & -0.35 & -0.10 \\
QLl & 2.07 & -0.26 & 0.34 & 0.46 & 0.83 & -0.40 & 0.30 \\
QLq & 2.91 & -1.13 & -0.22 & \textbf{-0.45} & 0.19 & \textbf{-1.42} & \textbf{-0.47} \\
VO & \textbf{1.83} & -0.38 & 0.22 & 0.49 & 0.05 & -0.36 & 0.15 \\
\bottomrule
\end{tabular}
\caption{Capability \gls{auc} change ($\Delta$) for gemma-3-4b in the one-minus-accuracy reward setting. Lowest values per condition are bold.}
\label{tab:gemma-3-4b-oma-ability-auc}
\end{table*}

\begin{table*}
\centering
\begin{tabular}{lrrrrrrr}
\toprule
 &  & \multicolumn{6}{c}{\textbf{Capability AUC $\Delta$ vs.\ Baseline}} \\
\cmidrule(lr){3-8}
 & \multicolumn{1}{l}{\textbf{Baseline}} & \textbf{QLl$=0$} & \textbf{QLq$=0$} & \multicolumn{2}{c}{\textbf{QLl$>$QLq}} & \multicolumn{2}{c}{\textbf{QLq$>$QLl}} \\
\cmidrule(lr){5-8}
 &  &  &  & \textbf{+1} & \textbf{+2} & \textbf{+1} & \textbf{+2} \\
\midrule
AS & 3.19 & 0.17 & -0.18 & 0.05 & 0.03 & -0.08 & 0.17 \\
AT & \textbf{3.74} & 0.23 & 0.29 & 0.35 & 0.12 & 0.03 & 0.07 \\
CEc & 3.51 & 0.17 & 0.52 & 0.71 & 0.46 & -0.29 & 0.04 \\
KNa & 3.25 & 0.10 & 0.10 & -0.14 & -0.00 & 0.15 & -0.02 \\
KNc & 3.57 & 0.13 & 0.09 & 0.15 & 0.11 & 0.09 & 0.04 \\
KNf & 3.45 & 0.03 & 0.52 & 0.57 & 0.16 & 0.07 & -0.05 \\
KNs & 2.99 & \textbf{0.51} & 0.14 & -0.03 & 0.16 & \textbf{0.22} & \textbf{0.55} \\
MCr & 2.77 & 0.09 & -0.13 & -0.20 & -0.02 & 0.04 & 0.20 \\
MCt & 3.23 & 0.03 & \textbf{0.65} & 0.80 & 0.18 & 0.04 & -0.06 \\
MCu & 2.88 & 0.22 & 0.19 & 0.30 & 0.05 & -0.03 & 0.15 \\
MS & 3.04 & 0.24 & 0.38 & 0.24 & 0.10 & 0.10 & 0.23 \\
QLl & 3.50 & 0.00 & 0.30 & 0.50 & 0.16 & -0.10 & -0.13 \\
QLq & 3.64 & 0.05 & 0.46 & \textbf{1.09} & \textbf{0.83} & -0.35 & 0.15 \\
VO & 3.15 & 0.08 & -0.26 & -0.03 & 0.03 & 0.03 & -0.06 \\
\bottomrule
\end{tabular}
\caption{Capability \gls{auc} change ($\Delta$) for gemma-3-12b in the accuracy reward setting. Highest values per condition are bold.}
\label{tab:gemma-3-12b-dw-ability-auc}
\end{table*}

\begin{table*}
\centering
\begin{tabular}{lrrrrrrr}
\toprule
 &  & \multicolumn{6}{c}{\textbf{Capability AUC $\Delta$ vs.\ Baseline}} \\
\cmidrule(lr){3-8}
 & \multicolumn{1}{l}{\textbf{Baseline}} & \textbf{QLl$=0$} & \textbf{QLq$=0$} & \multicolumn{2}{c}{\textbf{QLl$>$QLq}} & \multicolumn{2}{c}{\textbf{QLq$>$QLl}} \\
\cmidrule(lr){5-8}
 &  &  &  & \textbf{+1} & \textbf{+2} & \textbf{+1} & \textbf{+2} \\
\midrule
AS & 3.19 & 0.09 & -0.02 & -0.24 & -0.21 & -0.39 & -0.16 \\
AT & 3.74 & -0.06 & -0.13 & -0.26 & -0.09 & -0.45 & 0.21 \\
CEc & 3.51 & -1.24 & -1.33 & -1.27 & -0.17 & -1.62 & -0.91 \\
KNa & 3.25 & -1.46 & \textbf{-1.78} & -1.42 & -0.05 & -1.71 & -0.71 \\
KNc & 3.57 & -1.39 & -1.40 & -1.29 & -0.02 & -1.59 & -0.63 \\
KNf & 3.45 & -0.39 & -0.50 & -0.40 & \textbf{-0.35} & -1.02 & -0.24 \\
KNs & 2.99 & \textbf{-1.55} & -1.73 & \textbf{-1.55} & -0.07 & \textbf{-1.79} & -1.06 \\
MCr & \textbf{2.77} & -0.16 & -0.72 & -0.58 & -0.08 & -0.64 & -0.28 \\
MCt & 3.23 & -1.02 & -1.20 & -1.24 & -0.03 & -1.37 & -0.75 \\
MCu & 2.88 & -0.71 & -0.95 & -0.77 & -0.03 & -1.05 & -0.27 \\
MS & 3.04 & -0.96 & -1.74 & -0.94 & 0.09 & -1.00 & 0.38 \\
QLl & 3.50 & -0.59 & -0.64 & -0.63 & -0.25 & -0.48 & -0.67 \\
QLq & 3.64 & -0.98 & -1.08 & -1.16 & 0.14 & -1.52 & \textbf{-1.10} \\
VO & 3.15 & -0.29 & -0.54 & -0.82 & 0.09 & -1.20 & -0.38 \\
\bottomrule
\end{tabular}
\caption{Capability \gls{auc} change ($\Delta$) for gemma-3-12b in the one-minus-accuracy reward setting. Lowest values per condition are bold.}
\label{tab:gemma-3-12b-oma-ability-auc}
\end{table*}

\begin{table*}
\centering
\begin{tabular}{lrrrrrrr}
\toprule
 &  & \multicolumn{6}{c}{\textbf{Capability AUC $\Delta$ vs.\ Baseline}} \\
\cmidrule(lr){3-8}
 & \multicolumn{1}{l}{\textbf{Baseline}} & \textbf{QLl$=0$} & \textbf{QLq$=0$} & \multicolumn{2}{c}{\textbf{QLl$>$QLq}} & \multicolumn{2}{c}{\textbf{QLq$>$QLl}} \\
\cmidrule(lr){5-8}
 &  &  &  & \textbf{+1} & \textbf{+2} & \textbf{+1} & \textbf{+2} \\
\midrule
AS & 1.79 & 0.09 & 0.35 & 0.19 & 0.32 & 0.28 & 0.12 \\
AT & \textbf{2.22} & 0.40 & 0.47 & 0.27 & 0.46 & 0.53 & 0.40 \\
CEc & 1.24 & 0.82 & 0.61 & 0.72 & 0.71 & 1.01 & 0.62 \\
KNa & 1.23 & \textbf{1.47} & 1.42 & 1.11 & \textbf{1.21} & 1.29 & 1.31 \\
KNc & 1.55 & 1.25 & 1.11 & 0.90 & 0.84 & 1.25 & 1.09 \\
KNf & 1.29 & 0.25 & 0.53 & 0.25 & 0.32 & 0.62 & 0.31 \\
KNs & 1.44 & 1.30 & 1.22 & 1.14 & 1.04 & 1.10 & 1.23 \\
MCr & 1.41 & 0.27 & 0.54 & 0.35 & 0.52 & 0.66 & 0.55 \\
MCt & 1.35 & 0.27 & 0.44 & 0.33 & 0.31 & 0.77 & 0.29 \\
MCu & 1.37 & 0.48 & 0.75 & 0.63 & 0.47 & 0.52 & 0.71 \\
MS & 1.02 & 1.47 & \textbf{1.98} & \textbf{1.83} & 1.17 & \textbf{1.72} & \textbf{1.71} \\
QLl & 1.50 & -0.12 & 0.26 & -0.11 & 0.04 & 0.21 & -0.16 \\
QLq & 0.85 & 0.28 & 0.23 & 0.05 & 0.22 & 0.69 & 0.25 \\
VO & 1.08 & 0.15 & 0.39 & 0.24 & 0.20 & 0.33 & 0.17 \\
\bottomrule
\end{tabular}
\caption{Capability \gls{auc} change ($\Delta$) for Llama-3.2-3B in the accuracy reward setting. Highest values per condition are bold.}
\label{tab:llama-3.2-3b-dw-ability-auc}
\end{table*}

\begin{table*}
\centering
\begin{tabular}{lrrrrrrr}
\toprule
 &  & \multicolumn{6}{c}{\textbf{Capability AUC $\Delta$ vs.\ Baseline}} \\
\cmidrule(lr){3-8}
 & \multicolumn{1}{l}{\textbf{Baseline}} & \textbf{QLl$=0$} & \textbf{QLq$=0$} & \multicolumn{2}{c}{\textbf{QLl$>$QLq}} & \multicolumn{2}{c}{\textbf{QLq$>$QLl}} \\
\cmidrule(lr){5-8}
 &  &  &  & \textbf{+1} & \textbf{+2} & \textbf{+1} & \textbf{+2} \\
\midrule
AS & 1.79 & -1.01 & -0.86 & -1.30 & -1.78 & -1.19 & -0.99 \\
AT & 2.22 & \textbf{-1.37} & \textbf{-1.36} & \textbf{-1.63} & \textbf{-1.98} & \textbf{-1.54} & \textbf{-1.31} \\
CEc & 1.24 & -1.06 & -0.87 & -0.84 & -1.12 & -0.87 & -0.53 \\
KNa & 1.23 & -1.00 & -0.84 & -0.43 & -1.10 & -0.88 & -0.76 \\
KNc & 1.55 & -1.31 & -1.22 & -0.95 & -1.35 & -0.94 & -0.57 \\
KNf & 1.29 & -0.68 & -0.72 & -1.02 & -1.25 & -1.05 & -0.81 \\
KNs & 1.44 & -1.33 & -1.16 & -0.64 & -- & -0.63 & -0.55 \\
MCr & 1.41 & -0.91 & -0.97 & -1.15 & -- & -1.12 & -0.65 \\
MCt & 1.35 & -1.13 & -1.02 & -1.10 & -1.28 & -1.12 & -0.85 \\
MCu & 1.37 & -0.85 & -0.87 & -1.15 & -- & -1.01 & -0.91 \\
MS & 1.02 & -- & -0.89 & -0.77 & -- & -0.65 & -0.28 \\
QLl & 1.50 & -0.89 & -0.89 & -1.27 & -1.45 & -1.21 & -1.04 \\
QLq & \textbf{0.85} & -0.63 & -0.49 & -0.54 & -0.77 & -0.61 & -0.44 \\
VO & 1.08 & -0.83 & -0.73 & -0.76 & -1.04 & -0.69 & -0.38 \\
\bottomrule
\end{tabular}
\caption{Capability \gls{auc} change ($\Delta$) for Llama-3.2-3B in the one-minus-accuracy reward setting. Lowest values per condition are bold. ``--'' indicates missing values, e.g., where a fit was not possible.}
\label{tab:llama-3.2-3b-oma-ability-auc}
\end{table*}

\begin{table*}
\centering
\begin{tabular}{lrrrrrrr}
\toprule
 &  & \multicolumn{6}{c}{\textbf{Capability AUC $\Delta$ vs.\ Baseline}} \\
\cmidrule(lr){3-8}
 & \multicolumn{1}{l}{\textbf{Baseline}} & \textbf{QLl$=0$} & \textbf{QLq$=0$} & \multicolumn{2}{c}{\textbf{QLl$>$QLq}} & \multicolumn{2}{c}{\textbf{QLq$>$QLl}} \\
\cmidrule(lr){5-8}
 &  &  &  & \textbf{+1} & \textbf{+2} & \textbf{+1} & \textbf{+2} \\
\midrule
AS & 2.42 & 0.08 & 0.18 & 0.35 & -0.01 & 0.37 & 0.21 \\
AT & \textbf{2.95} & 0.15 & 0.36 & 0.75 & 0.37 & 0.67 & 0.51 \\
CEc & 1.82 & 0.30 & \textbf{1.09} & 1.13 & -0.16 & 0.68 & 1.09 \\
KNa & 1.62 & 0.78 & 0.96 & 0.80 & -0.20 & \textbf{1.34} & 0.89 \\
KNc & 2.23 & \textbf{0.80} & 0.90 & 1.11 & 0.18 & 1.29 & 0.99 \\
KNf & 2.07 & -0.07 & 0.30 & 0.76 & -0.22 & 0.32 & 0.21 \\
KNs & 2.15 & 0.70 & 0.85 & 1.21 & 0.29 & 0.84 & \textbf{1.18} \\
MCr & 1.84 & 0.25 & 0.34 & 0.51 & \textbf{0.39} & 0.83 & 0.46 \\
MCt & 1.60 & 0.26 & 0.84 & 0.76 & -0.00 & 0.62 & 0.69 \\
MCu & 1.89 & 0.33 & 0.86 & 1.15 & 0.32 & 0.69 & 0.75 \\
MS & 2.31 & 0.79 & 0.65 & 0.54 & 0.21 & 0.64 & 0.54 \\
QLl & 1.74 & 0.27 & 0.91 & \textbf{1.23} & 0.21 & 0.75 & 0.71 \\
QLq & 1.06 & 0.02 & 0.70 & 0.89 & 0.08 & 0.84 & 0.92 \\
VO & 1.23 & 0.05 & 0.35 & 0.26 & 0.15 & 0.77 & 0.46 \\
\bottomrule
\end{tabular}
\caption{Capability \gls{auc} change ($\Delta$) for Llama-3.1-8B in the accuracy reward setting. Highest values per condition are bold.}
\label{tab:llama-3.1-8b-dw-ability-auc}
\end{table*}

\begin{table*}
\centering
\begin{tabular}{lrrrrrrr}
\toprule
 &  & \multicolumn{6}{c}{\textbf{Capability AUC $\Delta$ vs.\ Baseline}} \\
\cmidrule(lr){3-8}
 & \multicolumn{1}{l}{\textbf{Baseline}} & \textbf{QLl$=0$} & \textbf{QLq$=0$} & \multicolumn{2}{c}{\textbf{QLl$>$QLq}} & \multicolumn{2}{c}{\textbf{QLq$>$QLl}} \\
\cmidrule(lr){5-8}
 &  &  &  & \textbf{+1} & \textbf{+2} & \textbf{+1} & \textbf{+2} \\
\midrule
AS & 2.42 & -0.87 & -0.77 & -2.27 & -2.12 & -1.67 & -1.70 \\
AT & 2.95 & -1.38 & -1.41 & \textbf{-2.76} & \textbf{-2.35} & \textbf{-1.88} & -1.84 \\
CEc & 1.82 & -1.19 & -1.25 & -1.56 & -1.48 & -1.26 & -1.13 \\
KNa & 1.62 & -1.34 & -1.42 & -1.57 & -1.49 & -1.32 & -0.97 \\
KNc & 2.23 & -1.74 & -1.74 & -2.02 & -1.74 & -1.09 & -1.00 \\
KNf & 2.07 & -0.58 & -0.68 & -1.89 & -1.71 & -1.63 & -1.57 \\
KNs & 2.15 & -2.11 & \textbf{-2.07} & -1.94 & -1.83 & -1.16 & -1.16 \\
MCr & 1.84 & -0.62 & -0.71 & -1.73 & -1.67 & -1.47 & -1.46 \\
MCt & 1.60 & -1.04 & -1.10 & -1.44 & -1.28 & -1.16 & -1.00 \\
MCu & 1.89 & -0.88 & -0.98 & -1.83 & -1.73 & -1.60 & -1.55 \\
MS & 2.31 & \textbf{-2.18} & -- & -- & -2.18 & -1.63 & \textbf{-1.90} \\
QLl & 1.74 & -0.42 & -0.56 & -1.62 & -1.46 & -1.46 & -1.29 \\
QLq & \textbf{1.06} & -0.79 & -0.73 & -0.91 & -0.79 & -0.62 & -0.47 \\
VO & 1.23 & -0.68 & -0.69 & -1.12 & -1.00 & -0.81 & -0.65 \\
\bottomrule
\end{tabular}
\caption{Capability \gls{auc} change ($\Delta$) for Llama-3.1-8B in the one-minus-accuracy reward setting. Lowest values per condition are bold. ``--'' indicates missing values, e.g., where a fit was not possible.}
\label{tab:llama-3.1-8b-oma-ability-auc}
\end{table*}

\begin{table*}
\centering
\begin{tabular}{lrrrrrrr}
\toprule
 &  & \multicolumn{6}{c}{\textbf{Capability AUC $\Delta$ vs.\ Baseline}} \\
\cmidrule(lr){3-8}
 & \multicolumn{1}{l}{\textbf{Baseline}} & \textbf{QLl$=0$} & \textbf{QLq$=0$} & \multicolumn{2}{c}{\textbf{QLl$>$QLq}} & \multicolumn{2}{c}{\textbf{QLq$>$QLl}} \\
\cmidrule(lr){5-8}
 &  &  &  & \textbf{+1} & \textbf{+2} & \textbf{+1} & \textbf{+2} \\
\midrule
AS & 2.80 & -0.05 & 0.08 & 0.20 & 0.20 & 0.07 & -0.02 \\
AT & \textbf{3.38} & 0.02 & 0.16 & 0.22 & 0.25 & 0.16 & -0.14 \\
CEc & 2.64 & \textbf{0.26} & 0.13 & \textbf{0.53} & 0.20 & 0.21 & \textbf{0.24} \\
KNa & 3.17 & -0.27 & -0.22 & -0.24 & -0.24 & -0.06 & -0.02 \\
KNc & 3.28 & 0.08 & -0.04 & 0.33 & \textbf{0.34} & 0.21 & 0.04 \\
KNf & 3.06 & -0.11 & -0.05 & 0.33 & 0.14 & 0.17 & 0.03 \\
KNs & 3.12 & -0.16 & -0.06 & -0.07 & -0.12 & 0.27 & -0.20 \\
MCr & 2.54 & -0.05 & 0.03 & 0.10 & 0.07 & 0.13 & -0.09 \\
MCt & 2.72 & -0.10 & -0.02 & 0.16 & 0.16 & 0.00 & 0.04 \\
MCu & 2.70 & -0.04 & \textbf{0.17} & 0.02 & -0.06 & 0.22 & -0.16 \\
MS & 3.09 & 0.06 & -0.19 & -0.20 & -0.14 & 0.06 & 0.00 \\
QLl & 2.92 & -0.20 & -0.17 & 0.07 & 0.04 & -0.13 & -0.16 \\
QLq & 2.11 & -0.01 & -0.05 & 0.08 & 0.15 & \textbf{0.29} & -0.02 \\
VO & 3.02 & -0.42 & -0.26 & -0.17 & -0.22 & -0.13 & -0.19 \\
\bottomrule
\end{tabular}
\caption{Capability \gls{auc} change ($\Delta$) for Qwen3-4B in the accuracy reward setting. Highest values per condition are bold.}
\label{tab:qwen3-4b-dw-ability-auc}
\end{table*}

\begin{table*}
\centering
\begin{tabular}{lrrrrrrr}
\toprule
 &  & \multicolumn{6}{c}{\textbf{Capability AUC $\Delta$ vs.\ Baseline}} \\
\cmidrule(lr){3-8}
 & \multicolumn{1}{l}{\textbf{Baseline}} & \textbf{QLl$=0$} & \textbf{QLq$=0$} & \multicolumn{2}{c}{\textbf{QLl$>$QLq}} & \multicolumn{2}{c}{\textbf{QLq$>$QLl}} \\
\cmidrule(lr){5-8}
 &  &  &  & \textbf{+1} & \textbf{+2} & \textbf{+1} & \textbf{+2} \\
\midrule
AS & 2.80 & -1.29 & -1.29 & -0.16 & -0.42 & -0.22 & -0.24 \\
AT & 3.38 & -1.67 & -1.71 & -0.29 & \textbf{-1.17} & \textbf{-0.39} & -0.30 \\
CEc & 2.64 & -1.66 & -0.68 & 0.00 & -0.04 & 0.17 & -0.01 \\
KNa & 3.17 & -2.61 & -1.52 & -0.40 & -0.49 & -0.14 & -0.19 \\
KNc & 3.28 & -2.40 & -1.62 & -0.23 & -0.55 & 0.23 & -0.16 \\
KNf & 3.06 & -1.48 & -1.12 & -0.01 & -0.44 & 0.13 & -0.21 \\
KNs & 3.12 & \textbf{-2.94} & -2.02 & -0.22 & -0.72 & -0.13 & -0.06 \\
MCr & 2.54 & -1.37 & -1.32 & -0.34 & -0.27 & -0.03 & -0.12 \\
MCt & 2.72 & -1.82 & -0.73 & -0.19 & -0.23 & -0.01 & -0.20 \\
MCu & 2.70 & -1.47 & -1.40 & -0.44 & -0.39 & -0.34 & -0.25 \\
MS & 3.09 & -- & \textbf{-2.09} & \textbf{-0.44} & -0.34 & -0.23 & -0.05 \\
QLl & 2.92 & -1.73 & -1.29 & -0.16 & -0.64 & -0.12 & \textbf{-0.34} \\
QLq & \textbf{2.11} & -0.95 & -0.20 & -0.04 & 0.05 & -0.05 & -0.10 \\
VO & 3.02 & -1.50 & -1.17 & -0.29 & -0.49 & -0.34 & -0.13 \\
\bottomrule
\end{tabular}
\caption{Capability \gls{auc} change ($\Delta$) for Qwen3-4B in the one-minus-accuracy reward setting. Lowest values per condition are bold. ``--'' indicates missing values, e.g., where a fit was not possible.}
\label{tab:qwen3-4b-oma-ability-auc}
\end{table*}

\begin{table*}
\centering
\begin{tabular}{lrrrrrrr}
\toprule
 &  & \multicolumn{6}{c}{\textbf{Capability AUC $\Delta$ vs.\ Baseline}} \\
\cmidrule(lr){3-8}
 & \multicolumn{1}{l}{\textbf{Baseline}} & \textbf{QLl$=0$} & \textbf{QLq$=0$} & \multicolumn{2}{c}{\textbf{QLl$>$QLq}} & \multicolumn{2}{c}{\textbf{QLq$>$QLl}} \\
\cmidrule(lr){5-8}
 &  &  &  & \textbf{+1} & \textbf{+2} & \textbf{+1} & \textbf{+2} \\
\midrule
AS & 3.03 & -0.08 & -0.01 & 0.04 & 0.03 & 0.03 & -0.03 \\
AT & 3.54 & 0.06 & 0.11 & 0.25 & 0.13 & 0.16 & 0.03 \\
CEc & 2.83 & 0.21 & 0.03 & \textbf{0.34} & 0.19 & 0.23 & \textbf{0.43} \\
KNa & 3.05 & -0.31 & -0.19 & -0.32 & 0.05 & -0.04 & -0.20 \\
KNc & \textbf{3.58} & -0.03 & -0.02 & 0.13 & 0.22 & 0.12 & -0.02 \\
KNf & 3.47 & \textbf{0.31} & 0.24 & 0.26 & 0.37 & 0.14 & 0.01 \\
KNs & 2.97 & -0.14 & 0.08 & 0.16 & 0.17 & 0.17 & 0.09 \\
MCr & 2.77 & -0.09 & -0.15 & 0.06 & -0.03 & -0.06 & 0.09 \\
MCt & 2.94 & 0.12 & -0.18 & 0.01 & 0.06 & 0.05 & 0.04 \\
MCu & 2.68 & 0.15 & \textbf{0.41} & 0.21 & 0.19 & 0.12 & 0.17 \\
MS & 2.80 & -0.25 & -0.09 & 0.02 & \textbf{0.51} & 0.13 & 0.07 \\
QLl & 3.17 & -0.16 & 0.10 & 0.11 & 0.13 & -0.12 & -0.02 \\
QLq & 2.30 & -0.12 & -0.08 & 0.13 & 0.03 & \textbf{0.26} & 0.03 \\
VO & 2.83 & -0.32 & -0.33 & -0.01 & -0.00 & 0.10 & 0.01 \\
\bottomrule
\end{tabular}
\caption{Capability \gls{auc} change ($\Delta$) for Qwen3-8B in the accuracy reward setting. Highest values per condition are bold.}
\label{tab:qwen3-8b-dw-ability-auc}
\end{table*}

\begin{table*}
\centering
\begin{tabular}{lrrrrrrr}
\toprule
 &  & \multicolumn{6}{c}{\textbf{Capability AUC $\Delta$ vs.\ Baseline}} \\
\cmidrule(lr){3-8}
 & \multicolumn{1}{l}{\textbf{Baseline}} & \textbf{QLl$=0$} & \textbf{QLq$=0$} & \multicolumn{2}{c}{\textbf{QLl$>$QLq}} & \multicolumn{2}{c}{\textbf{QLq$>$QLl}} \\
\cmidrule(lr){5-8}
 &  &  &  & \textbf{+1} & \textbf{+2} & \textbf{+1} & \textbf{+2} \\
\midrule
AS & 3.03 & -1.16 & -0.26 & -0.20 & -1.05 & -0.27 & -0.34 \\
AT & 3.54 & -1.83 & \textbf{-0.55} & -0.26 & \textbf{-1.96} & -0.31 & -0.39 \\
CEc & 2.83 & -1.70 & 0.05 & 0.14 & -0.46 & -0.48 & -0.32 \\
KNa & 3.05 & -2.41 & -0.17 & -0.26 & -0.25 & -0.17 & -0.19 \\
KNc & 3.58 & -2.78 & -0.42 & -0.11 & -0.91 & -0.04 & -0.01 \\
KNf & 3.47 & -1.36 & -0.04 & -0.07 & -1.19 & -0.53 & -0.65 \\
KNs & 2.97 & \textbf{-2.80} & -0.28 & -0.17 & -0.71 & 0.11 & 0.06 \\
MCr & 2.77 & -1.59 & -0.30 & -0.10 & -0.88 & -0.13 & -0.21 \\
MCt & 2.94 & -1.91 & -0.13 & \textbf{-0.27} & -0.61 & -0.56 & -0.59 \\
MCu & 2.68 & -1.50 & -0.31 & -0.18 & -0.85 & 0.08 & -0.29 \\
MS & 2.80 & -2.34 & -0.18 & -0.26 & -0.12 & 0.10 & -0.07 \\
QLl & 3.17 & -1.55 & 0.00 & 0.04 & -1.13 & -0.44 & -0.25 \\
QLq & \textbf{2.30} & -1.15 & 0.09 & 0.07 & -0.10 & -0.44 & -0.38 \\
VO & 2.83 & -1.39 & 0.07 & 0.07 & -0.50 & \textbf{-0.78} & \textbf{-0.80} \\
\bottomrule
\end{tabular}
\caption{Capability \gls{auc} change ($\Delta$) for Qwen3-8B in the one-minus-accuracy reward setting. Lowest values per condition are bold.}
\label{tab:qwen3-8b-oma-ability-auc}
\end{table*}

\begin{table*}[!ht]
\centering
\setlength{\tabcolsep}{3.5pt}
\small
\begin{tabular}{lllrrrrrc}
\toprule
\multirow{2}{*}{\textbf{Reward}} & \multirow{2}{*}{\textbf{Train}} & \multirow{2}{*}{\textbf{Test}} &
\multicolumn{3}{c}{\textbf{Baseline}} & \multicolumn{2}{c}{\textbf{$\Delta$ (pp)}} & \multirow{2}{*}{\textbf{Imp./Worse}} \\
\cmidrule(lr){4-6} \cmidrule(lr){7-8}
& & & Min & Median & Max & Mean & Median & \\
\midrule

\multirow[t]{16}{*}{Acc.}
& \multirow[t]{4}{*}{TRAM-Ari.} 
& TRAM-Ari. & 25.74 & 70.09 & 87.74 & 23.12 & 14.65 & 6/0 \\
& & TRAM-Sem. & 10.00 & 61.40 & 70.60 & 11.13 &  1.90 & 5/0 \\
& & ToT-Ari.  & 15.37 & 50.22 & 72.08 &  0.83 &  1.73 & 4/2 \\
& & ToT-Sem.  & 27.86 & 42.50 & 55.86 &  0.86 &  1.07 & 4/2 \\
\cmidrule(lr){2-9}

& \multirow[t]{4}{*}{TRAM-Sem.}
& TRAM-Ari. & 25.74 & 70.09 & 87.74 &  3.35 &  1.52 & 3/3 \\
& & TRAM-Sem. & 10.00 & 61.40 & 70.60 & 15.17 &  5.00 & 6/0 \\
& & ToT-Ari.  & 15.37 & 50.22 & 72.08 & -0.72 & -0.43 & 3/3 \\
& & ToT-Sem.  & 27.86 & 42.50 & 55.86 &  1.12 &  0.43 & 3/3 \\
\cmidrule(lr){2-9}

& \multirow[t]{4}{*}{ToT-Ari.}
& TRAM-Ari. & 25.74 & 70.09 & 87.74 & -3.22 & -2.31 & 2/4 \\
& & TRAM-Sem. & 10.00 & 61.40 & 70.60 & -2.87 &  0.40 & 4/2 \\
& & ToT-Ari.  & 15.37 & 50.22 & 72.08 &  7.61 &  7.79 & 6/0 \\
& & ToT-Sem.  & 27.86 & 42.50 & 55.86 &  1.40 &  0.86 & 3/3 \\
\cmidrule(lr){2-9}

& \multirow[t]{4}{*}{ToT-Sem.}
& TRAM-Ari. & 25.74 & 70.09 & 87.74 & -1.01 & -0.85 & 2/4 \\
& & TRAM-Sem. & 10.00 & 61.40 & 70.60 & -0.00 &  0.60 & 4/2 \\
& & ToT-Ari.  & 15.37 & 50.22 & 72.08 &  1.01 &  1.19 & 5/1 \\
& & ToT-Sem.  & 27.86 & 42.50 & 55.86 &  6.81 &  7.00 & 6/0 \\
\midrule

\multirow[t]{16}{*}{1-Acc.}
& \multirow[t]{4}{*}{TRAM-Ari.}
& TRAM-Ari. & 25.74 & 70.09 & 87.74 & -17.17 & -16.67 & 2/4 \\
& & TRAM-Sem. & 10.00 & 61.40 & 70.60 &   0.80 &   0.40 & 3/3 \\
& & ToT-Ari.  & 15.37 & 50.22 & 72.08 &  -5.77 &  -4.76 & 1/5 \\
& & ToT-Sem.  & 27.86 & 42.50 & 55.86 &  -1.62 &  -0.79 & 3/3 \\
\cmidrule(lr){2-9}

& \multirow[t]{4}{*}{TRAM-Sem.}
& TRAM-Ari. & 25.74 & 70.09 & 87.74 &  -1.22 &  -0.78 & 2/4 \\
& & TRAM-Sem. & 10.00 & 61.40 & 70.60 & -12.93 & -11.50 & 1/5 \\
& & ToT-Ari.  & 15.37 & 50.22 & 72.08 &  -1.12 &  -0.22 & 2/4 \\
& & ToT-Sem.  & 27.86 & 42.50 & 55.86 &  -6.36 &  -2.29 & 1/5 \\
\cmidrule(lr){2-9}

& \multirow[t]{4}{*}{ToT-Ari.}
& TRAM-Ari. & 25.74 & 70.09 & 87.74 & -10.07 &  -6.06 & 1/5 \\
& & TRAM-Sem. & 10.00 & 61.40 & 70.60 &  -6.03 &  -1.60 & 0/6 \\
& & ToT-Ari.  & 15.37 & 50.22 & 72.08 & -25.97 & -18.51 & 0/6 \\
& & ToT-Sem.  & 27.86 & 42.50 & 55.86 & -12.31 & -10.21 & 1/5 \\
\cmidrule(lr){2-9}

& \multirow[t]{4}{*}{ToT-Sem.}
& TRAM-Ari. & 25.74 & 70.09 & 87.74 &  -0.95 &  -1.33 & 2/4 \\
& & TRAM-Sem. & 10.00 & 61.40 & 70.60 &  -0.60 &  -1.30 & 1/5 \\
& & ToT-Ari.  & 15.37 & 50.22 & 72.08 &  -2.63 &  -1.73 & 1/5 \\
& & ToT-Sem.  & 27.86 & 42.50 & 55.86 & -22.76 & -25.00 & 0/6 \\
\bottomrule
\end{tabular}

\vspace{2pt}
\footnotesize
TRAM-Ari. = TRAM Arithmetic, TRAM-Sem. = TRAM Semantic, ToT-Ari. = ToT Arithmetic, ToT-Sem. = ToT Semantic.
\caption{Model performance deltas relative to baseline (percentage points) for both reward settings. Includes the number of models which improved or worsened their performance after fine-tuning.}\label{tab:model-performance-deltas}
\end{table*}

\section{Non-temporal QL-heavy training}\label{app:ql-heavy}

To test whether a stronger quantitative and logical reasoning signal transfers to temporal reasoning, we fine-tune all six models on the non-temporal QL-dominant data described in Section~\ref{sec:ql-heavy-training}. Table~\ref{tab:non-temporal-ql-fam-capability-auc} reports the mean capability
\gls{auc} change per dimension, averaged across models, and~\ref{tab:non-temporal-ql-fam-temporal-mix-acc} the per-model accuracy on the temporal mixture before and after this training. The training achieves its
intended effect on the targeted capabilities: \gls{qll} and \gls{qlq} show the largest gains of all 14 dimensions ($+0.15$ \gls{auc} each). Accuracy on the temporal mixture, however, is essentially unchanged for every model (mean $\Delta=-0.06$~pp; per-model range $-1.23$ to $+0.55$~pp), indicating that increased QL capability alone does not translate into better performance on the evaluated temporal benchmarks (see Section~\ref{sec:drivers}).

\begin{table*}
\centering
\begin{tabular}{lr}
\toprule
\textbf{Dimension} & $\boldsymbol{\Delta}$ AUC \\
\midrule
AS & +0.01 \\
AT & -0.02 \\
CEc & +0.14 \\
KNa & -0.04 \\
KNc & +0.02 \\
KNf & +0.08 \\
KNs & +0.07 \\
MCr & +0.03 \\
MCt & +0.02 \\
MCu & -0.02 \\
MS & +0.07 \\
QLl & \textbf{+0.15} \\
QLq & \textbf{+0.15} \\
VO & -0.08 \\
\bottomrule
\end{tabular}
\caption{Mean capability \gls{auc} $\Delta$ versus baseline after GRPO on non-temporal QL-dominant data, averaged across models. Highest values are bold.}
\label{tab:non-temporal-ql-fam-capability-auc}
\end{table*}

\begin{table*}
\centering
\begin{tabular}{lrrr}
\toprule
 & \textbf{Baseline} & \multicolumn{2}{c}{\textbf{QL-dominant.}} \\
\cmidrule(lr){2-2}\cmidrule(lr){3-4}
 & \textbf{Acc.} & \textbf{Acc.} & $\boldsymbol{\Delta}$ \\
\midrule
Llama-3.2-3B & 24.85 & 25.40 & +0.55 \\
Llama-3.1-8B & 33.33 & 32.87 & -0.46 \\
gemma-3-4b & 43.42 & 43.71 & +0.29 \\
gemma-3-12b & 62.53 & 62.78 & +0.25 \\
Qwen3-4b & 53.59 & 52.36 & -1.23 \\
Qwen3-8b & 55.78 & 56.03 & +0.25 \\
\bottomrule
\end{tabular}
\caption{Overall accuracy on the temporal mix (\%) after GRPO on non-temporal QL-dominant training data. $\Delta$ is percentage points versus each model's baseline.}
\label{tab:non-temporal-ql-fam-temporal-mix-acc}
\end{table*}

\section{Larger models}
\label{app:larger_models}

\begin{figure*}[t]
    \centering
    \import{figures/}{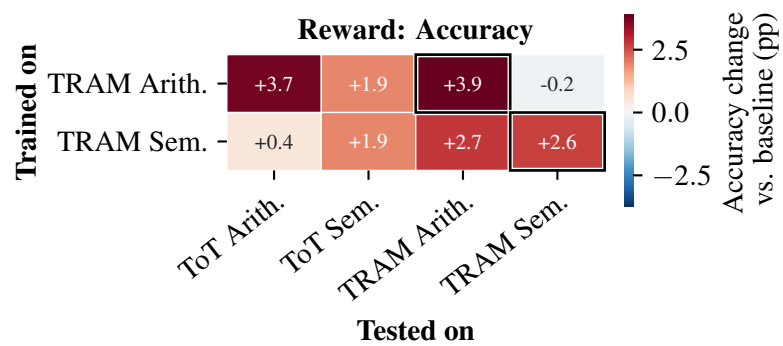}
    \caption{Change in accuracy, in percentage points, between the untrained Qwen3-32B baseline and each trained variant.}
    \label{fig:accuracy_delta_matrix_qwen3_32b}
\end{figure*}

We also explored scaling the experiments to larger models, but the computational cost of GRPO prevented us from running the full experimental suite. For example, the length of some \gls{tot} prompts, which can span several thousand tokens, would have been much higher than for TRAM. Nevertheless, we trained Qwen3-32B separately on the TRAM Arithmetic and Semantic subsets using the standard accuracy-reward configuration and evaluated both variants on all TRAM and \gls{tot} subsets. Figure~\ref{fig:accuracy_delta_matrix_qwen3_32b} reports the resulting changes in accuracy.

Training on TRAM Arithmetic improves \gls{tot} Arithmetic by 3.7 percentage points, nearly matching the 3.9-point in-domain gain, while \gls{tot} Semantic improves by 1.9 points. Training on TRAM Semantic improves \gls{tot} Semantic by 1.9 points, but also increases TRAM Arithmetic by 2.7 points. Overall, however, the changes range from $-0.2$ to $+3.9$ points, substantially smaller than the in-domain gains of $+6.8$ to $+23.1$ points observed across the models in Figure~\ref{fig:accuracy_delta_matrix}. These results therefore provide tentative evidence of arithmetic transfer, but not a consistent mode-specific transfer pattern. Because they come from a single larger model, it remains unclear whether the observed pattern reflects model scale or ordinary performance variation.

\end{document}